\documentclass[acmsmall]{acmart}
\newtheorem{definition}{Definition}
\AtBeginDocument{%
  }

\setcopyright{cc}
\setcctype{by}
\acmDOI{10.1145/3798212}
\acmYear{2026}
\acmJournal{PACMPL}
\acmVolume{10}
\acmNumber{OOPSLA1}
\acmArticle{104}
\acmMonth{4}
\received{2025-10-10}
\received[accepted]{2026-02-17}

\usepackage{multirow}
\usepackage{algorithm}
\usepackage{algpseudocode}
\usepackage{amsmath}
\usepackage{subcaption}
\usepackage{tcolorbox}
\usepackage{float}
\usepackage[capitalise]{cleveref}
\usepackage{pifont}
\usepackage{bbding}
\usepackage{wrapfig}
\usepackage{xcolor}
\usepackage{graphicx}
\usepackage{enumitem}
\usepackage{booktabs}
\usepackage{siunitx}
\newcommand{\m}{\mathit}

\newcommand{\drule}{Q}
 
\newcommand{\fact}{F}

\newcommand{\hornarrow}{\,\text{:--}\,}
\newcommand{\head}[1]{{\noindent\textbf{#1}}}

\definecolor{customorange}{HTML}{F8CDAC}
\definecolor{customgreen}{HTML}{C6E2B1}

\definecolor{ForestGreen}{RGB}{34,139,34}

\crefformat{section}{\S~#2#1#3}
\crefformat{subsection}{\S#2#1#3}
\crefformat{subsubsection}{\S#2#1#3}

\crefmultiformat{section}{\S\S#2#1#3}{ and~#2#1#3}{, #2#1#3}{, and~#2#1#3}

\crefrangeformat{section}{\S\S#3#1#4 to~#5#2#6}

\Crefname{section}{Section}{Sections}

\begin{document}

\title{MetaSpace: Metamorphic Testing for Spatial Cognition in Embodied Agents}

\author{Gengyang Xu}
\email{gxuah@cse.ust.hk}
\orcid{0009-0001-7221-7845}
\affiliation{%
  \institution{Hong Kong University of Science and Technology}
  \city{Hong Kong}
  \country{China}
}

\author{Dongwei Xiao}
\email{dxiaoad@cse.ust.hk}
\orcid{0000-0002-4680-5715}
\affiliation{%
  \institution{Hong Kong University of Science and Technology}
  \city{Hong Kong}
  \country{China}}
\authornote{Corresponding authors.}

\author{Yiteng Peng}
\email{ypengbp@cse.ust.hk}
\orcid{0009-0006-2066-7939}
\affiliation{%
  \institution{Hong Kong University of Science and Technology}
  \city{Hong Kong}
  \country{China}}
\authornotemark[1]

\author{Shuai Wang}
\email{shuaiw@cse.ust.hk}
\orcid{0000-0002-0866-0308}
\affiliation{%
 \institution{Hong Kong University of Science and Technology}
  \city{Hong Kong}
 \country{China}}

\renewcommand{\shortauthors}{Gengyang Xu, Dongwei Xiao, Yiteng Peng, and Shuai Wang}

\begin{abstract}
An embodied agent is an intelligent entity that interacts with its
environment through a physical body. Currently, the evaluation of embodied
agents primarily relies on two paradigms: (1) manually annotated Visual
Question Answering (VQA) pairs and (2) high-level task completion metrics, such as
success in navigation or manipulation. The former is labor-intensive and subject
to variability in annotation quality. The latter may obscure critical
vulnerabilities, allowing agents to complete tasks through suboptimal means or
safety violations, thereby concealing safety risks and inefficiencies. Given
that spatial cognition is the cornerstone for executing embodied tasks, there is
a pressing need to assess whether embodied agents possess robust
spatial cognition during task execution. 

Inspired by metamorphic testing principles in software engineering, we propose
MetaSpace, a novel framework designed to evaluate the spatial cognition of
agents. By leveraging spatiotemporal multimodal states derived from real
execution trajectories, MetaSpace automatically generates test cases based on
predefined metamorphic relations (MRs) grounded in logical rules and physical
laws. Crucially, we encode these MRs as executable rules in a logic programming
language (Prolog). Violations of these relations indicate failures in spatial
cognition. Our empirical evaluation across three embodied scenarios
demonstrates that MetaSpace successfully detects 90,422 spatial cognition errors
in state-of-the-art (SOTA) MLLM-driven agents. We introduce the Spatial Cognition (SC)
score to quantify performance. Results indicate that all SOTA
agents achieve average scores between 0.44 and 0.52, significantly lower than
the human benchmark of 0.96. Additionally, these agents struggle with
directional tasks, with SC scores consistently below 0.38. In contrast, their
performance in magnitude-related tasks is relatively better, with most SC scores
exceeding 0.5. To mitigate the identified spatial cognition errors, we explore
potential improvement strategies. Preliminary results suggest that traditional
prompting techniques (e.g., Chain of Thought) are limited, while spatially-aware
prompting (e.g., cognitive maps) shows promise. Our findings underscore the
importance of ongoing community efforts to enhance embodied agent performance by
prioritizing the improvement of spatial cognition, a fundamental requirement for
executing embodied tasks.
\end{abstract}

\begin{CCSXML}
<ccs2012>
<concept>
<concept_id>10011007.10011074.10011099.10011102.10011103</concept_id>
<concept_desc>Software and its engineering~Software testing and debugging</concept_desc>
<concept_significance>500</concept_significance>
</concept>
</ccs2012>
\end{CCSXML}



\maketitle

\section{Introduction}
Embodied agents are attracting significant attention from both academia and industry, with applications spanning robotics~\cite{roy2021,yang2025embodiedbench,li2024embodied}, autonomous driving~\cite{tian2024drivevlm}, and drones~\cite{zhao2025urbanvideo,gao2024embodiedcity}. Embodied agents are capable of performing a wide range of tasks, from semantic tasks (e.g., household chores) to core embodied functions (e.g., navigation, rearrangement, and manipulation). 
Leveraging Multi-modal Large Language Models (MLLMs) to create embodied agents presents a promising avenue for tackling embodied tasks~\cite{li2024embodied,yang2025embodiedbench}. However, while Large Language Models (LLMs) have achieved remarkable success in linguistic tasks~\cite{rostam2024achieving,10.1145/3744746,10433480,10.1145/3728902}, they face critical challenges in visuospatial tasks, which significantly undermine their effectiveness and reliability in embodied intelligence applications.

To evaluate the performance of embodied agents, the standard paradigm in the AI and robotics community relies on benchmarking with manually annotated Visual Question Answering (VQA) pairs (e.g., Multiple-Choice Questions (MCQs))~\cite{du-etal-2024-embspatial,ramakrishnan2025doesspatialcognitionemerge,ma20253dsrbenchcomprehensive3dspatial,zhao2025urbanvideo,cheng2025embodiedeval,gao2024embodiedcity,Dang_2025_CVPR}. However, the manual design and annotation of test cases are labor-intensive, and variability in annotator expertise can introduce inconsistency and bias into benchmark assessments. Additionally, some benchmarks (e.g., ~\cite{ramakrishnan2025doesspatialcognitionemerge}) utilize classic cognitive psychology questions originally designed for humans or animals, such as the Minnesota Paper Form Board (MPFB) test~\cite{likert1941minnesota}. These approaches fail to capture the essence of \emph{embodiment} in agents, leading to inadequate reflection of their actual performance in real-world applications.

\begin{figure}[t]
    \centering
    \begin{subfigure}{0.32\textwidth}
        \centering
        \includegraphics[width=\linewidth]{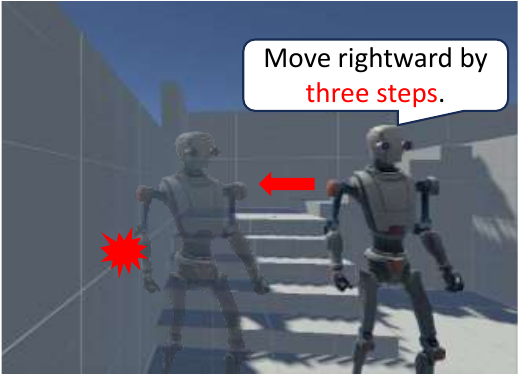}
        \caption{Wrong Magnitude Perception}
        \label{fig:false_positive_success_1}
    \end{subfigure}
    \hfill
    \begin{subfigure}{0.32\textwidth}
        \centering
        \includegraphics[width=\linewidth]{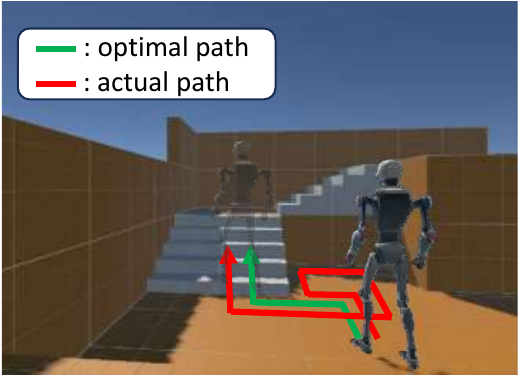}
        \caption{Redundant Trial-and-Error}
        \label{fig:false_positive_success_2}
    \end{subfigure}
    \hfill
    \begin{subfigure}{0.32\textwidth}
        \centering
        \includegraphics[width=\linewidth]{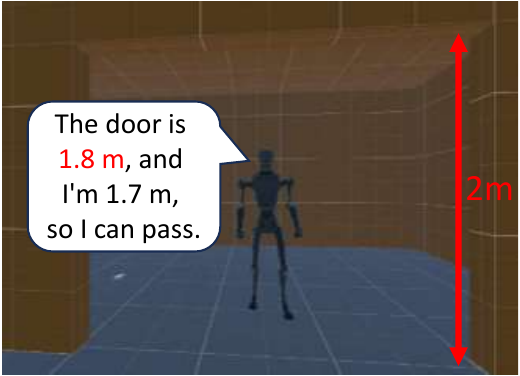}
        \caption{Coincidence-Driven Success}
        \label{fig:false_positive_success_3}
    \end{subfigure}
    \caption{Examples of ``false positive success'' in high-level embodied tasks.}
    \label{fig:false_positive_success}
\end{figure}

In contrast, some studies have begun to rely on high-level task completion
metrics, such as whether an agent reaches navigation targets or successfully
performs manipulation
tasks~\cite{yang2025embodiedbench,choi2024lota,zhou2024hazard,cheng2025embodiedeval}.\footnote{In
this context, we define low-level tasks as the foundational tasks that underpin
high-level tasks (e.g., navigation and manipulation). For instance, successful
high-level navigation relies on low-level reasoning about object directional
relations to avoid obstacles.} While intuitive and easy to quantify, these
outcome-oriented evaluations conceal underlying flaws, and agents may succeed in
completing tasks through non-optimal means or safety violations (``false
positive success''). For example, (1) \textit{Wrong magnitude perception}: as
shown in Fig.~\ref{fig:false_positive_success_1}, agents may complete tasks
based on erroneous action magnitude perception (e.g., taking three steps for a
navigation move, when only two are necessary). In virtual environments, the
absence of physical collisions allows agents to create an illusion of success.
(2) \textit{Redundant trial-and-error}: agents may misjudge spatial information
(e.g., direction perception errors) and correct their trajectories through
repeated adjustments, thereby sacrificing efficiency for task completion without
resulting in task failure (Fig.~\ref{fig:false_positive_success_2}). (3)
\textit{Coincidence-driven success}: spatial cognition errors may not always
result in failure, as agents might, by chance, avoid negative consequences
(e.g., in Fig.~\ref{fig:false_positive_success_3}, the agent misjudges a 2-meter
door as \SI{1.8}{meters} but successfully passes through without getting stuck due to
its own height of \SI{1.7}{meters}). These ``false positive successes'' are not rare;
existing embodied benchmark studies have similar observations through manual
checking~\cite{li2024embodied,yang2025embodiedbench,cheng2025embodiedeval}. Such
defects in the existing evaluation paradigm conceal significant safety risks
(e.g., non-catastrophic collisions) and operational inefficiencies (e.g.,
suboptimal paths), undermining the trustworthiness of embodied agents in real-world
applications.

The high-level task outcome-oriented evaluation paradigm is an end-to-end
approach that assesses only the final results of task execution, without
examining the intermediate processes or decision-making involved. In reality,
embodied task execution is inherently compositional: successful completion of
high-level tasks relies on first accomplishing a series of spatial cognitive
sub-tasks. For example, to navigate to a target location, an agent must
understand the spatial relationships between the target and surrounding
landmarks, and accurately perceive its own movement direction and distance to
ensure correct actions. In other words, spatial cognition (e.g., movement
perception, spatial awareness) serves as the cornerstone for embodied agents to
complete high-level embodied tasks. These observations highlight our key
motivation for this research:
\begin{quote}
  ``\textit{Is spatial cognition, as the keystone of high-level embodied tasks,
  truly a solved problem for embodied agents?}''
\end{quote}

To bridge the identified research gap, it is essential to move beyond outcome-oriented testing paradigms and adopt a capability-oriented approach that enables an authentic evaluation of embodied capabilities. Developing such a capability-oriented evaluation framework requires us to address three significant challenges:
\textbf{\textit{C1: Inadequacy of Static Evaluation for Embodiment.}} Existing methods for spatial cognition evaluations often fail to capture the dynamic essence of embodiment. Most rely on static Visual Question Answering (VQA) tasks, which lack interactive engagement with the environment and do not assess the agent's ability to transform between egocentric and allocentric perspectives.
\textbf{\textit{C2: Challenge in Defining Test Oracle.}} Automatically determining the ground truth of spatial cognition is inherently challenging. Embodied tasks are dynamic and context-dependent, making it difficult to define the expected spatial relations and perceptions for all possible scenarios.
\textbf{\textit{C3: Challenge in Test Case Generation.}} Manual design and annotation of test cases continues to be the prevailing practice. However, this approach is both labor-intensive and prone to incompleteness, frequently missing rare or edge cases 
for robust testing. Furthermore, the quality and consistency of benchmark questions can vary significantly based on the expertise of human annotators, introducing noise and potential bias into the assessment.

To address the above challenges, we propose MetaSpace, a metamorphic testing
(MT)~\cite{chen2020metamorphic} framework specifically designed to evaluate the
spatial cognition of embodied agents. MetaSpace shifts the focus of embodied
agent assessment from external outcome to internal capability, specifically
targeting SC abilities. Importantly, all test cases are automatically generated
from real embodied task execution trajectories, ensuring a strong correlation
between evaluation results and actual embodied performance. To address
\textbf{C1} and \textbf{C3}, MetaSpace collects spatiotemporal multimodal states
derived from real trajectories and utilizes these states to generate test cases.
To address \textbf{C2}, we craft a set of metamorphic relations (MRs) to serve
as test oracles; these MRs are based on principled logical rules (e.g.,
transitivity, symmetry) and physical laws (e.g., perspective geometry).
We use logic programming to encode these MRs as executable rules and agents' spatial observations as facts to automate the validation process. MetaSpace provides a comprehensive assessment of eight key embodied spatial cognitive abilities, ensuring a robust evaluation of embodied agents in real-world embodied scenarios. In summary, our contributions are threefold:
\begin{itemize}
\item \textbf{At the conceptual level}, we move beyond simple end-to-end outcomes and instead scrutinize the intrinsic spatial cognitive abilities of embodied agents. This capability-based view unlocks transparency into the decision process itself, which is the foundation for high-level embodied tasks. We identify eight key spatial cognitive abilities essential for embodied agents, drawing insights from cognitive psychology and embodied intelligence literature.

\item \textbf{At the technical level}, we develop MetaSpace, a MT framework that implements a set of logic- and physics-based MRs to evaluate embodied spatial cognition. These MRs are encoded into a Prolog knowledge base, allowing for scalable and oracle-free validation.

\item \textbf{At the empirical level}, we apply MetaSpace to test six state-of-the-art (SOTA) MLLM-driven embodied agents across three real-world scenarios. MetaSpace executes 30,300 unique test cases on six agents, uncovering a total of 90,422 instances that trigger spatial cognitive errors. Based on our findings, we derive valuable insights and offer recommendations for mitigating spatial cognition errors in embodied agents. Preliminary experiments demonstrate that cognitive map prompting can enhance agents' directional spatial cognitive abilities.
\end{itemize}

\section{Background}
\subsection{Embodied Spatial Cognition}
\label{sec:embodied_spatial_cognition}

\head{Definition.}
The concept of spatial cognition originates from cognitive psychology. It refers
to the study of knowledge and beliefs regarding the spatial properties of
objects and events, including aspects such as location, size, distance, and
movement~\cite{MONTELLO200114771}. We particularly focus on \textit{embodied
spatial cognition}, which pertains to the spatial cognition of embodied
intelligence in supporting embodied tasks.\footnote{In this paper, we use
``spatial cognition'', ``SC'', and ``spatial cognitive capability''
interchangeably to describe ``embodied spatial cognition''.}

\head{Scope.}
While humans can derive spatial cognition from various modalities (e.g., one can estimate the location of a ringing phone even without seeing it), most embodied agents primarily rely on visual input at this stage~\cite{burgess2008spatial}. Given this context, our work focuses on visual-spatial cognition. Additionally, while cognitive psychology provides classic spatial cognition experiments, including pen-and-paper tasks and the Minnesota Paper Form Board Test (MPFB)~\cite{likert1941minnesota}, our emphasis is on spatial cognition within embodied scenarios, which involves dynamic interactions with the environment.

\begin{figure}[t]
    \centering
    \includegraphics[width=\textwidth]{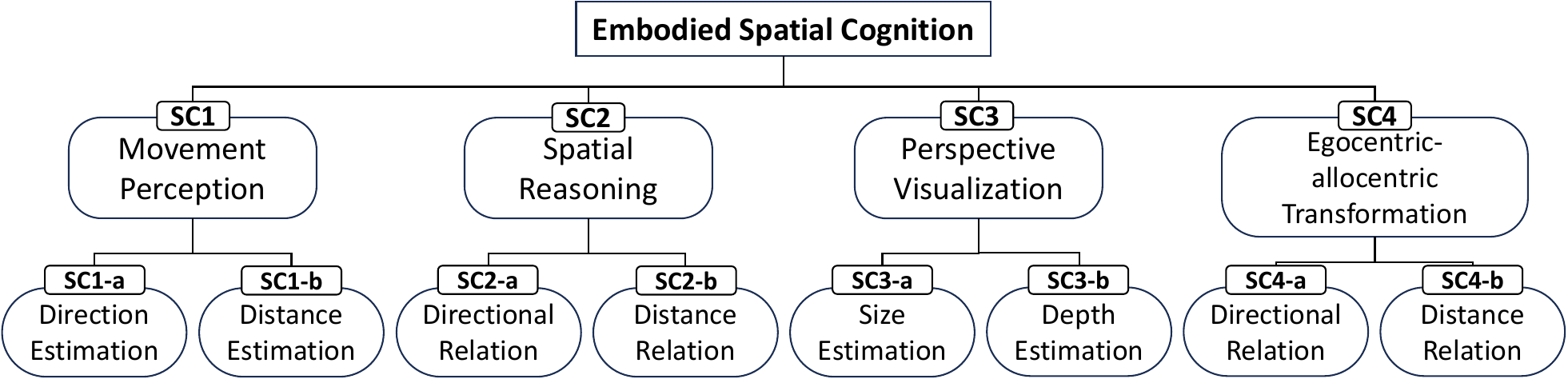}
    \caption{Taxonomy of embodied spatial cognition with numbering codes.}
    \label{fig:taxonomy}
\end{figure}

\head{Taxonomy.}
We present a taxonomy of capabilities essential for embodied spatial cognition (Fig.~\ref{fig:taxonomy}). Rather than an arbitrary collection, our taxonomy synthesizes foundational domains from cognitive science and psychology. Specifically, we identify four key aspects, each grounded in established theories:
\textit{SC1: Movement perception.} Grounded in Gibson's ecological theory of perception~\cite{gibson1979ecological}, this capability captures the agent's ability to sense self-motion for immediate control. It is foundational for basic navigation and manipulation tasks.
\textit{SC2: Spatial reasoning.} Drawing on Kosslyn's theory of spatial relations~\cite{kosslyn1987seeing}, we assess the ability to identify spatial relationships between objects and landmarks. 
Furthermore, based on Burgess's model of spatial memory~\cite{burgess2006spatial}, \textit{SC4: egocentric-allocentric transformation} assesses the critical ability to translate between egocentric views and allocentric mental maps. This is vital for envisioning actions from future viewpoints.
\textit{SC3: Perspective visualization.} This is essential as embodied agents operate in 3D environments. Aligned with Marr's computational vision theory~\cite{marr2010vision}, we assess the recovery of intrinsic properties (i.e., size and depth) from 2D observations (e.g., the 2.5D sketch). 
Across SC1, SC2, and SC4, the distinction between directional (SC*-a) and magnitude-based (SC*-b) capabilities is supported by Kosslyn's theory of spatial relations~\cite{kosslyn1987seeing}, which differentiates between categorical spatial processing (e.g., relative directions) and coordinate spatial processing (e.g., precise distances). This suggests they involve distinct cognitive mechanisms, warranting separate evaluation. In contrast, SC3 focuses on recovering intrinsic properties (size and depth) from 2D observations, consistent with Marr's vision theory~\cite{marr2010vision}.

\subsection{Logic Programming for Spatial Cognition Validation}
\label{sec:logic_programming}
In this study, we apply logic programming to implement the designed MRs for evaluating embodied spatial cognition. In the context of embodied spatial cognition, we encode spatial observations as facts and MRs as rules, facilitating rigorous validation of the cognitive consistency of embodied agents without human intervention. We elaborate on the specific process below.

\head{Automatic Fact Generation.}
Agent responses regarding spatial observations are automatically converted into Prolog facts.~\footnote{MetaSpace constrains agents to respond in predefined formats through structured prompting. We discuss the influence of structured prompting in \cref{fig:case_study}.} For example, when an agent perceives movement from state \( s_1 \) to \( s_2 \) as ``going forward'', this generates the fact \( \m{moveForward(s_1, s_2).} \) Similarly, the object relationships observed by agents can be encoded into facts such as \( \m{eastOf(landmark_A, landmark_B).} \)

\head{Metamorphic Relation Rules.}
A rule is a conditional statement that allows new facts to be inferred from existing ones. A
common structure for a rule is the Horn clause, which comprises a head predicate and a rule body
(a list of predicates).
In MetaSpace, each MR is encoded as Horn clause rules defining consistency constraints. 
An example demonstrating the transitivity rule is $\m{eastOf(X,Z)\hornarrow eastOf(X,Y), eastOf(Y,Z)}$. This rule means that if X is east of Y, and Y is east of Z, the system can infer that X is east of Z.
Another example, $\m{westOf(X,Y) \hornarrow eastOf(Y,X)}$, defines $\m{westOf}$ as an inverse relation of $\m{eastOf}$.

\head{Automated MR Validation Program.}
A logic program, or knowledge base, consists of a collection of facts ($\widetilde{\fact}$) and a collection of rules ($\widetilde{\drule}$) that define the system's knowledge.
For each test case, MetaSpace constructs a logic program combining observed facts with MR rules:
\begin{equation}
\begin{aligned}
\m{(Program)} & \quad \mathcal{P}_{\text{spatial}} &{ ::= } & \quad \widetilde{\fact}_{\text{observations}} \,{+}{+}\, \widetilde{\drule}_{\text{MRs}}
\end{aligned}
\label{eq:program}
\end{equation}
Here, the tilde notation indicates a list of items. The Prolog engine then validates consistency by checking whether agent responses satisfy the expected spatial relations derived from the rules. Any inconsistency indicates a spatial cognition violation.

\section{Metamorphic Testing}
\subsection{Motivation for Using MT}

A fundamental challenge in evaluating embodied agents is the \textit{test oracle problem}, i.e., the difficulty of determining the correct, expected output for a given test case. Embodied tasks occur in dynamic environments, requiring real-time route planning as environmental conditions evolve. Therefore, evaluating the correctness and optimality of answers generated by embodied agents presents a significant challenge. Existing research often assesses embodied intelligence through MCQ benchmarks. These benchmarks typically consist of a set of images accompanied by manually designed questions, usually derived from human-generated scenarios. During the evaluation process, the agent is presented with the image, the corresponding question, and a selection of multiple-choice answers. However, the consistency and the quality of benchmark questions can vary significantly due to the differing levels of expertise among the human experts involved in their creation, which is not only labor-intensive but also prone to inconsistencies. Moreover, human-generated test cases may not adequately cover edge cases, as it is inherently challenging for human experts to annotate answers for tasks involving constantly changing scenarios.

MT offers a powerful solution to the test oracle problem and the limitations of
manual test generation in dynamic environments. Inspired by its significant
success in assessing the quality of
software~\cite{mansur2021metamorphic,10.1145/3293882.3330567}, AI
models~\cite{wang2020metamorphic}, compilers~\cite{xiao2022metamorphic,xiao2025mtzk}, and quantum computing
platforms~\cite{paltenghi2023morphq}, we adapt MT to evaluate the embodied
spatial cognition of embodied intelligence. The core strength of MT is its
ability to validate outputs without a predefined oracle. Instead of asserting
the exact output, it uses MRs to check for expected
consistencies. For instance, to test the implementation of \( \sin(x) \), we do
not need to know the expected output for arbitrary floating-point inputs \( x
\). Instead, we can assert that the \( \text{MR}: \sin(x) = \sin(\pi - x) \)
must always hold. A discrepancy between the outputs for \( \sin(x) \) and \(
\sin(\pi - x) \) reveals an implementation error, effectively bypassing the
oracle problem.

\subsection{Formulation of Conducting MT with MetaSpace}

We formalize the MT process in MetaSpace as follows. In MT, a
MR defines a predictable relationship between the outputs of an embodied agent
when its inputs are mutated. Specifically, for a function \( f: \mathcal{I} \to
\mathcal{O} \), where \( \mathcal{I} \) denotes the inputs (e.g., visual
observations), \( \mathcal{O} \) represents the outputs (e.g., perceived spatial
relations, movement estimations), and \( f \) is the embodied agent under test.
Based on a given MR, we perform a transformation $ \phi: \mathcal{I} \to
\mathcal{I} $ on an original input \( \mathbf{x} \in \mathcal{I} \) to generate
a new input \( \mathbf{x}' = \phi(\mathbf{x}) \in \mathcal{I} \). The
corresponding outputs are \( \mathbf{y} = f(\mathbf{x}) \) and \( \mathbf{y}' =
f(\mathbf{x}') \).

The MR is encoded as a boolean predicate \( \mathrm{MR}(\mathbf{x}, \mathbf{x}', \mathbf{y}, \mathbf{y}') \), which evaluates to \( \mathrm{True} \) if and only if \( \mathbf{y} \) and \( \mathbf{y}' \) satisfy a predefined consistency constraint \( \mathcal{P}_\phi(\mathbf{y}, \mathbf{y}') \) based on principled logical rules and physical laws:
\begin{equation}
\mathrm{MR}(\mathbf{x}, \mathbf{x}', \mathbf{y}, \mathbf{y}') = 
\begin{cases} 
\mathrm{True} & \iff \mathcal{P}_\phi(\mathbf{y}, \mathbf{y}'), \\
\mathrm{False} & \text{otherwise}.
\end{cases}
\label{eq:MR}
\end{equation}
Overall, the general process of conducting MT using MetaSpace involves the following steps:
\begin{enumerate}
  \item Select an initial input \( \mathbf{x} \in \mathcal{I} \) and obtain the output \( \mathbf{y} = f(\mathbf{x}) \).
  \item Apply a metamorphic transformation \( \phi \) compatible with \( \mathbf{x} \) to generate \( \mathbf{x}' = \phi(\mathbf{x}) \in \mathcal{I} \)), and obtain output \( \mathbf{y}' = f(\mathbf{x}') \).
  \item Check whether the tuple \( (\mathbf{x}, \mathbf{x}', \mathbf{y}, \mathbf{y}') \) satisfies the predefined MR. If not, it indicates a potential error in the agent's spatial cognitive responses.
\end{enumerate}

\subsection{MRs in MetaSpace}

MetaSpace implements MRs derived from principled logic rules and physical laws
to find potential errors in the outputs of embodied intelligence models. We
employ logic programming to implement these MRs (discussed soon in
\cref{sec:automated_validation}). These MRs holistically capture
diverse spatial cognitive capabilities in embodied scenarios mentioned in
\cref{fig:taxonomy}. Additionally, as we will demonstrate in
\textsection{\ref{sec:ablation_study}}, these MRs are highly accurate in
identifying spatial cognition errors in embodied agents.

\begin{table}
\small
\centering
\caption{MRs for Spatial Cognition Evaluation (Please refer to
Fig.~\ref{fig:taxonomy} for the spatial cognitive capabilities associated with
the codes in the table).}
\label{tab:MRs}
\begin{tabular}{clll}
\toprule
\textbf{Category} & \textbf{Metamorphic Relation} & \textbf{Foundational Principle} & \textbf{Spatial Capability} \\
\midrule
\multirow{3}{*}{Logical}  
& MR1: Transitivity         & Spatial Relation Logic        & SC1-a, SC2-a, SC4 \\
& MR2: Symmetry             & Spatial Relation Logic        & SC1, SC2, SC4     \\
& MR3: Contradiction           & Law of Non-Contradiction & SC1, SC2, SC4     \\
\midrule
\multirow{3}{*}{Physical} 
& MR4: Triangle Inequality  & Triangle Inequality      & SC1-b, SC2-b      \\
& MR5: Size-Depth Consistency       & Perspective Geometry     & SC3-b               \\
& MR6: Object Size Ratio Consistency & Perspective Geometry     & SC3-a               \\
\bottomrule
\end{tabular}
\end{table}

Our MRs can be broadly categorized into two groups based on their foundational
principles: logical consistency-oriented MRs
(\cref{sec:logical_MRs}) and physical law-oriented MRs
(\cref{sec:physical_MRs}). Table~\ref{tab:MRs} summarizes the six
MRs implemented in MetaSpace, along with their foundational principles and the
specific spatial cognitive capabilities they assess (as defined in
Fig.~\ref{fig:taxonomy}). In \cref{sec:logical_MRs} and
\cref{sec:physical_MRs}, we provide a detailed introduction to the
MRs designed in MetaSpace.

\subsubsection{Logical Consistency-Oriented MRs}
\label{sec:logical_MRs}
MRs based on logical consistency are designed to ensure that the outputs of
original and transformed test cases adhere to predefined logical rules. These
MRs are grounded in fundamental principles of logic, and they help identify
inconsistencies in the spatial cognitive responses of embodied agents.

\begin{itemize}
  \item \textbf{MR1: Transitivity.} This MR is grounded in the transitivity property of directional spatial relations. It asserts that any violation of the transitive property defined by Definition \ref{def:mr1_transitivity} indicates a potential spatial cognition error regarding directional relationships.

  \begin{definition}[MR1: Transitivity]
      \label{def:mr1_transitivity}
      Let $s_1, s_2, s_3$ be three states. In MetaSpace, $s_i$ can represent either (i) the spatial state of an object, where $Dir(s_i, s_j)$ denotes the spatial (e.g., directional) relation between objects' states $s_i$ and $s_j$ (e.g., north, east, southwest, above), or (ii) the state of an agent, where $Dir(s_i, s_j)$ denotes the agent's perceived movement direction from state $s_i$ to $s_j$ (e.g., go forward, backward, upward).
      The operator $\oplus$ denotes the composition of such relations.
      Given three state pairs $(s_1, s_2)$, $(s_2, s_3)$, and $(s_1, s_3)$, the relations among these pairs must satisfy the transitive property, defined as:
      \begin{equation}
        MR_1(s_1, s_2, s_3):\quad Dir(s_1, s_3) = Dir(s_1, s_2) \oplus Dir(s_2, s_3)
      \end{equation}
  That is, the direction or motion from \( s_1 \) to \( s_3 \) must equal the composition of the one from \( s_1 \) to \( s_2 \) and from \( s_2 \) to \( s_3 \), ensuring logical consistency.
      \end{definition}
Examples illustrating MR1 are shown in Fig.~\ref{fig:MR1}.
\textbf{Example 1:} With this MR, for \( s \) representing an object's state, if the directional relation between \( s_1 \) and \( s_2 \) is east, and between \( s_2 \) and \( s_3 \) is north, then the relation between \( s_1 \) and \( s_3 \) must be northeast (i.e., east \( \oplus \) north \( = \) northeast). 
\textbf{Example 2:} Similarly, if the directional relation between \( s_1 \) and \( s_2 \) is northeast, and between \( s_2 \) and \( s_3 \) is north, then the relation between \( s_1 \) and \( s_3 \) must still be northeast (i.e., northeast \( \oplus \) north \( = \) northeast). 
\textbf{Example 3:} Moreover, for \( s \) representing an agent's state, if an agent perceives the movement from \( s_1 \) to \( s_2 \) as moving forward, and from \( s_2 \) to \( s_3 \) as moving right, then the agent's response to the movement from \( s_1 \) to \( s_3 \) must be equivalent to the combination of these two movements (i.e., move forward \( \oplus \) move right).

    \begin{figure}[t]
        \centering
        \begin{subfigure}[b]{0.3\textwidth}
            \centering
            \includegraphics[width=\textwidth]{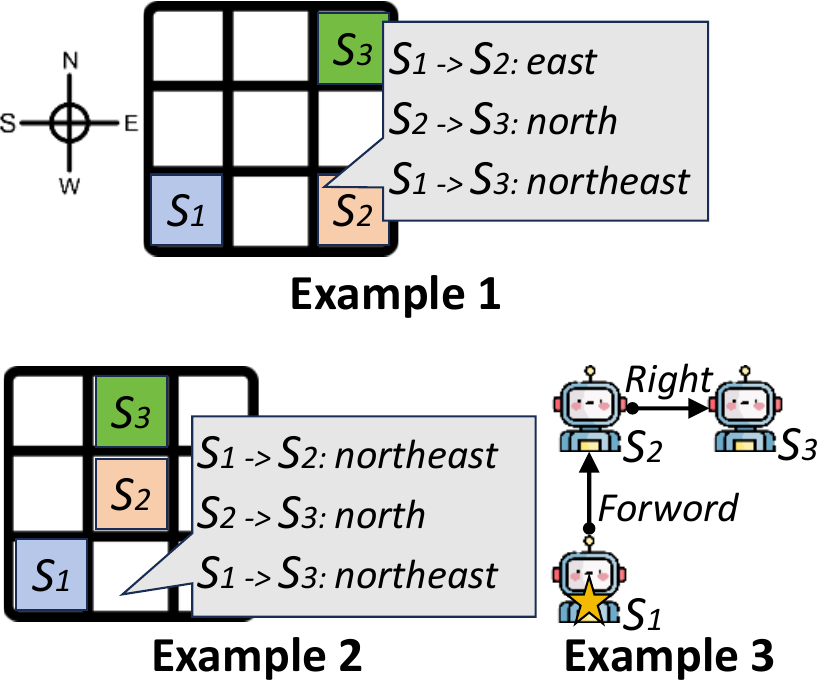}
            \caption{MR1: Transitivity}
            \label{fig:MR1}
        \end{subfigure}
        \hspace{0.05\textwidth}
        \begin{subfigure}[b]{0.3\textwidth}
            \centering
            \includegraphics[width=\textwidth]{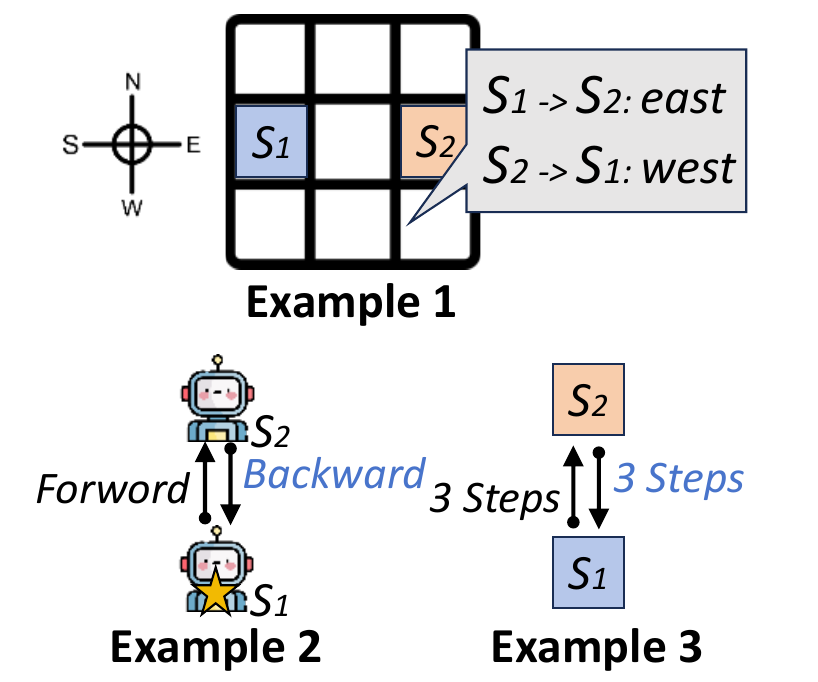}
            \caption{MR2: Symmetry}
            \label{fig:MR2}
        \end{subfigure}
        \caption{Examples of MR1 and MR2 in both object spatial relations and agent movement perceptions.}
        \label{fig:MR1_MR2}
    \end{figure}

  \item \textbf{MR2: Symmetry.} MR2 leverages the symmetry property in spatial relations. The directional relation or motion should satisfy the antisymmetry property, while the magnitude (distance) relation should satisfy the symmetry property, as defined in Definition \ref{def:mr2_symmetry}.

  \begin{definition}[MR2: Symmetry]
      \label{def:mr2_symmetry}
      Let $Dist(s_1, s_2)$ denote the distance from $s_1$ to $s_2$, and $Inv$ be the inverse operator for directional relations or movement directions. Given a state pair $(s_1, s_2)$, the direction or motion from $s_2$ to $s_1$ must be the inverse of the one from $s_1$ to $s_2$, and the distance from $s_2$ to $s_1$ must equal the distance from $s_1$ to $s_2$, as defined:
      \begin{equation}
        MR_2(s_1, s_2):\quad 
        \begin{cases}
          Dir(s_2, s_1) = Inv(Dir(s_1, s_2)) \\
          Dist(s_2, s_1) = Dist(s_1, s_2)
        \end{cases}
      \end{equation}
    \end{definition}
  We show examples illustrating MR2 in Fig.~\ref{fig:MR2}.
  \textbf{Example 1:} If the direction between \( s_1 \) and \( s_2 \) is east, then the direction between \( s_2 \) and \( s_1 \) must be west (i.e., \( Inv(\text{east}) = \text{west} \)). \textbf{Example 2:} Similarly, if the motion from \( s_1 \) to \( s_2 \) is moving forward, then the motion from \( s_2 \) to \( s_1 \) must be moving backward (i.e., \( Inv(\text{move forward}) = \text{move backward} \)).
  \textbf{Example 3:} If the distance between \( s_1 \) and \( s_2 \) is 3 steps, then the distance between \( s_2 \) and \( s_1 \) must also be 3 steps (i.e., \( Dist(s_2, s_1) = Dist(s_1, s_2) = 3 \) steps).\footnote{MetaSpace employs steps as distance units, as they are more relevant to embodied contexts.}

  \item \textbf{MR3: Contradiction.} MR3 is based on the law of non-contradiction, which states that contradictory statements cannot be true simultaneously, as defined in Definition \ref{def:mr3_contradiction}. It applies to both directional and magnitude spatial relationships.

  \begin{definition}[MR3: Contradiction]
      \label{def:mr3_contradiction}
      It is impossible for $s_1$ to be in two different directions from $s_2$ at
      the same time, and similarly, the distance between $s_1$ and $s_2$ cannot
      simultaneously be two different values. The MR is formally defined as:
        \begin{equation}
        MR_3(s_1, s_2):\quad
        \begin{cases}
          \neg \left( Dir(s_1, s_2) = d_1 \wedge Dir(s_1, s_2) = d_2 \right),\quad \forall d_1 \neq d_2 \\
          \neg \left( Dist(s_1, s_2) = m_1 \wedge Dist(s_1, s_2) = m_2 \right),\quad \forall m_1 \neq m_2
        \end{cases}
        \end{equation}
  \end{definition}

  \textbf{Example 1:} The agent cannot consider the motion from \( s_1 \) to \( s_2 \) as both moving forward and moving backward. In other words, it cannot justify that both answers are correct (i.e., \( \neg(\text{move forward} \wedge \text{move backward}) \)).
  \textbf{Example 2:} The agent cannot simultaneously justify that the direction from \( s_1 \) to \( s_2 \) is both east and west (i.e., \( \neg(\text{east} \wedge \text{west}) \)). 
  \textbf{Example 3:} Furthermore, the agent cannot perceive the distance from \( s_1 \) to \( s_2 \) as both 2 steps and 3 steps (i.e., \( \neg(2 \text{ steps} \wedge 3 \text{ steps}) \)). We illustrate MR3 in Fig.~\ref{fig:MR3}.
 
\end{itemize}

    \begin{figure}[t]
        \centering
        \begin{subfigure}[b]{\textwidth}
            \centering
            \includegraphics[width=0.8\textwidth]{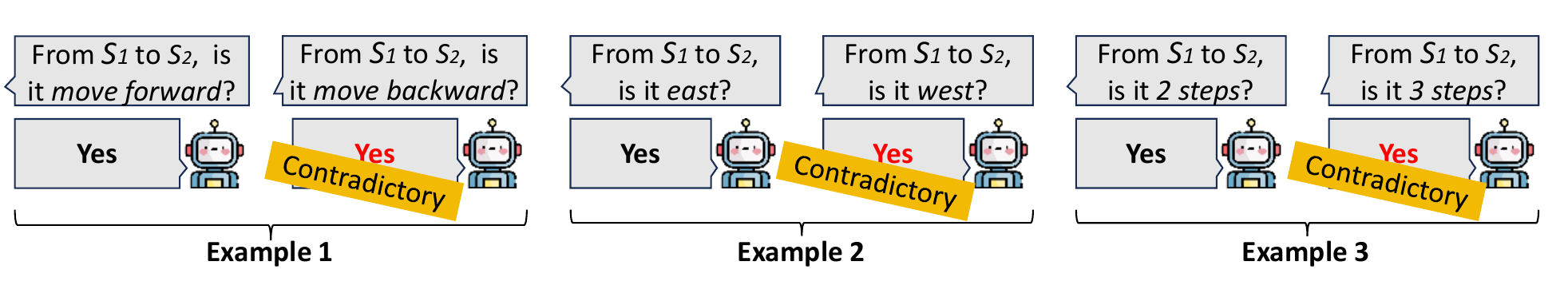}
            \caption{MR3: Contradiction}
            \label{fig:MR3}
        \end{subfigure}

        \begin{subfigure}[b]{\textwidth}
            \centering
            \includegraphics[width=0.75\textwidth]{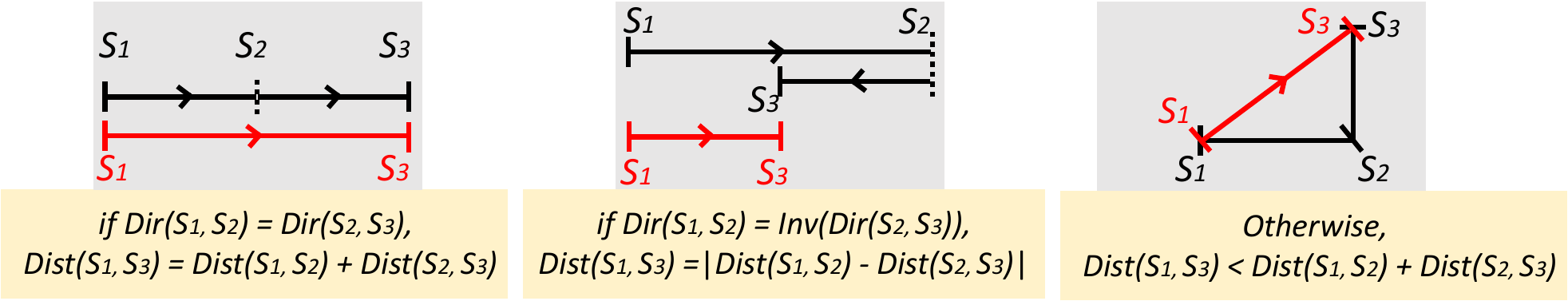}
            \caption{MR4: Triangle Inequality}
            \label{fig:MR4}
        \end{subfigure}
        \caption{Examples of MR3: Contradiction and MR4: Triangle Inequality.}
        \label{fig:MR3_MR4}
    \end{figure}

\subsubsection{Physical Law-Oriented MRs}
\label{sec:physical_MRs}
Physical law-oriented MRs validate whether the spatial cognition remains consistent with physical laws. They are defined as follows:

\begin{itemize}
    \item \textbf{MR4: Triangle Inequality.} MR4 is based on the triangle inequality theorem. The theorem states that in Euclidean space, for any triangle, the sum of the lengths of any two sides must be greater than or equal to the length of the remaining side. Any violation of this theorem indicates a potential spatial cognition error by the agent, as defined in Definition \ref{def:mr4_triangle_inequality}. We illustrate MR4 in \cref{fig:MR4}.
    
    \begin{definition}[MR4: Triangle Inequality]
      \label{def:mr4_triangle_inequality}
  Given three states $s_1, s_2, s_3$, if they are collinear (i.e., lie on the same straight line), then whether considering the spatial distance between objects or the movement distance of an agent, the distance between $s_1$ and $s_3$ equals either the sum or the absolute difference of the other two distances, depending on their relative directions. Otherwise, the distance between $s_1$ and $s_3$ must be strictly less than the sum of the other two distances.
      \begin{equation}
      MR_4(s_1, s_2, s_3):
      \begin{cases}
        Dist(s_1, s_3) = Dist(s_1, s_2) + Dist(s_2, s_3), & \text{if } Dir(s_1, s_2) = Dir(s_2, s_3) \\
        Dist(s_1, s_3) = \left|Dist(s_1, s_2) - Dist(s_2, s_3)\right|, & \text{if } Dir(s_1, s_2) = Inv(Dir(s_2, s_3)) \\
        Dist(s_1, s_3) < Dist(s_1, s_2) + Dist(s_2, s_3), & \text{otherwise}
      \end{cases}
      \end{equation}
    \end{definition}

    \item \textbf{MR5: Size-Depth Consistency.} This relation is grounded in the principles of perspective geometry and is used to validate the agent's perception of depth.
    
    \begin{definition}[MR5: Size-Depth Consistency]
      \label{def:mr5_size_depth_consistency}
      Let $o$ be a specific object observed by the agent in two different frames, $f_1$ and $f_2$.
      Let $S(f_i, o)$ denote the projected size (e.g., pixel height) of object $o$ in frame $f_i$, and $D(f_i, o)$ denote the depth (distance from the agent) of $o$ in frame $f_i$. Under the assumption of a pinhole camera model, perspective geometry gives $S(f, o) \propto \frac{1}{D(f, o)}$. Thus, for two frames:
      \begin{equation}
      MR_5(f_1, f_2, o):\quad \left| \frac{S(f_1, o)}{S(f_2, o)} - \frac{D(f_2, o)}{D(f_1, o)} \right| < \epsilon
      \end{equation}
      That is, the ratio of projected sizes should be inversely proportional to the ratio of depths, within a relative error tolerance $\epsilon$.
    \end{definition}
\cref{fig:MR5} illustrates this concept. In this MR, we adopt the ideal pinhole camera model with consistent intrinsic parameters (e.g., focal length, principal point) as a standard approximation. Although deviations may arise due to factors such as lens distortion, this assumption remains reasonable. This is because current MLLM-driven embodied intelligence primarily relies on RGB images as input and focuses on coarse-grained depth estimation during embodied tasks rather than precise depth estimation~\cite{yang2025embodiedbench,li2024embodied}. Therefore, we accept controllable deviations. 
Unlike the previous MRs, MR5 uses meters as the estimation unit. To accommodate acceptable deviations in depth estimation, we introduce a relative tolerance threshold \( \epsilon \). We discuss the threshold determination and sensitivity analysis in \textsection{\ref{sec:threshold_determination}}.

    \begin{figure}[t]
        \centering
        \begin{subfigure}[b]{0.3\textwidth}
            \centering
            \includegraphics[width=\textwidth]{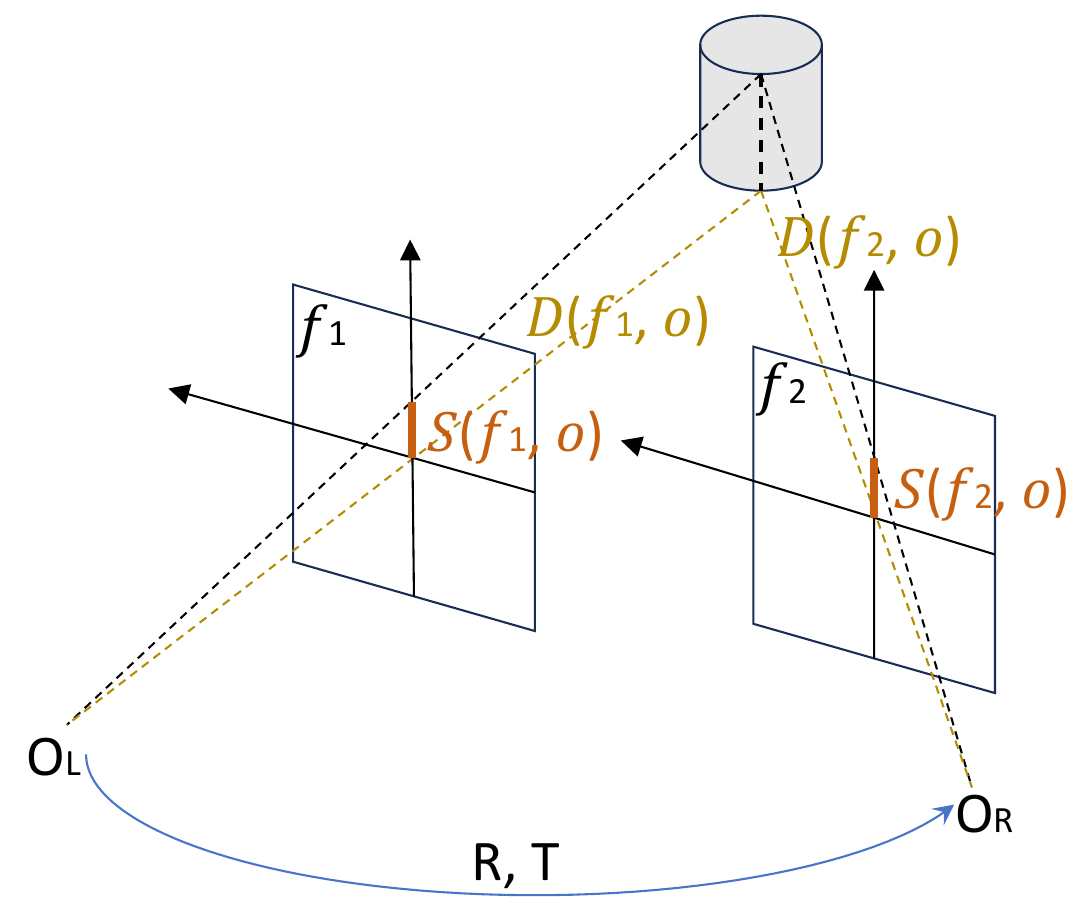}
            \caption{MR5}
            \label{fig:MR5}
        \end{subfigure}
        \hspace{0.05\textwidth}
        \begin{subfigure}[b]{0.3\textwidth}
            \centering
            \includegraphics[width=\textwidth]{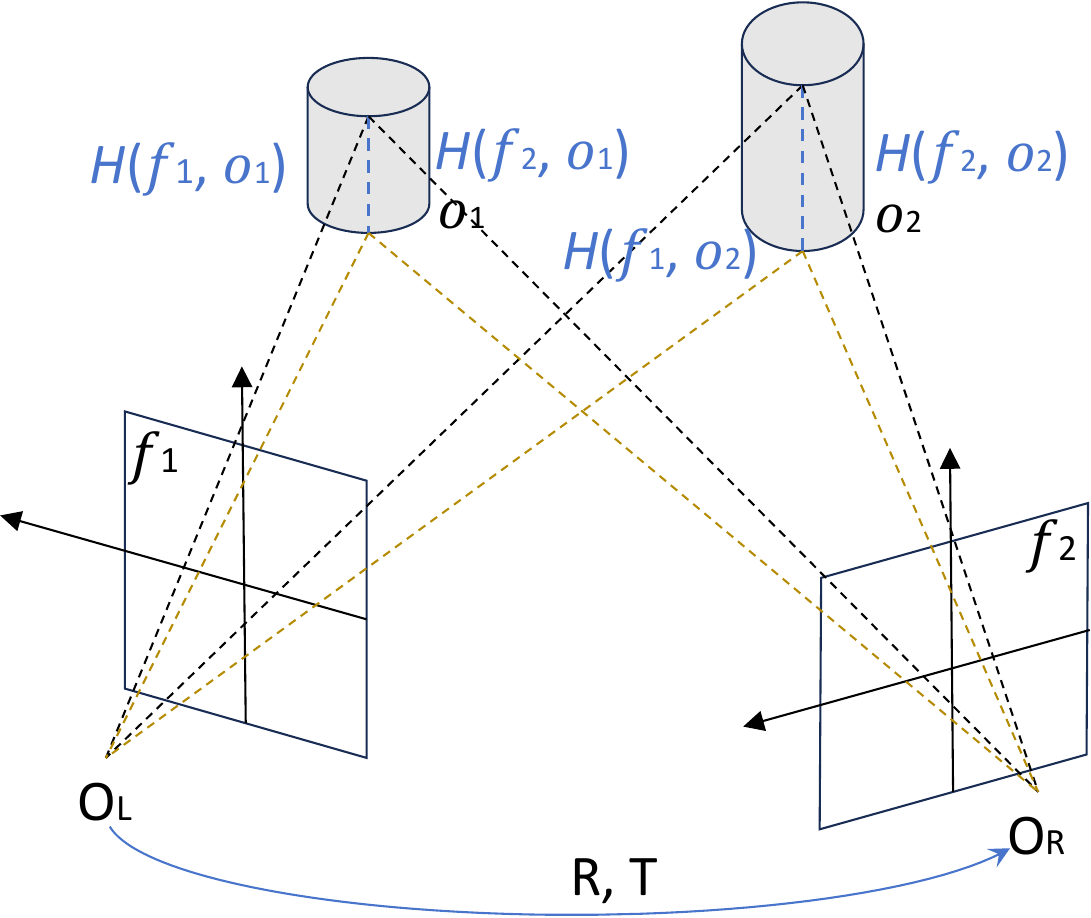}
            \caption{MR6}
            \label{fig:MR6}
        \end{subfigure}
        \caption{
            Under the pinhole camera model, perspective geometry dictates that the ratio of projected sizes should be inversely proportional to the ratio of depths for a single object. MR5 (Size-Distance Consistency for Multi-Frame Single Object) asserts that a single object's size-depth relationship remains consistent across different frames, while MR6 (Size-Distance Consistency for Multi-Object Multiple Frames) ensures that the size ratio between different objects remains consistent across frames.
        }
        \label{fig:mainfigure}
    \end{figure}

  \item \textbf{MR6: Object Size Ratio Consistency.} This relation is used to validate the agent's perception of object size, as defined in Definition \ref{def:mr6_object_size_ratio}.

  \begin{definition}[MR6: Object Size Ratio Consistency]
    \label{def:mr6_object_size_ratio}
          Let $f_1$ and $f_2$ be two frames from the agent's trajectory, and $o_1$ and $o_2$ be two distinct objects observed in both frames. Denote $H(f_i,o_j)$ as the estimated real-world size of object $o_j$ in frame $f_i$. The following approximate consistency relation should hold:
          \begin{equation}
            MR_6(f_1, f_2, o_1, o_2):\quad \left| \frac{H(f_1,o_1)}{H(f_1,o_2)} - \frac{H(f_2,o_1)}{H(f_2,o_2)} \right| < \delta
          \end{equation}
          That is, the ratio of the estimated real-world sizes of two objects should remain consistent across different frames within a relative error tolerance $\delta$. We discuss the determination and sensitivity analysis of this threshold in \textsection{\ref{sec:threshold_determination}}.
          The reason for comparing a ratio instead of absolute size is that we observe the scale of the virtual simulator environment is often altered by the stretching or scaling of 3D models.
      \end{definition}
\end{itemize}

\subsection{Discussions on MR Design}
\label{sec:discussion_on_MR_design}

Our set of six MRs is not an arbitrary collection of heuristics; rather, it is
derived through a systematic selection process governed by three rigorous
criteria: (1) \textbf{Theoretical Foundation.} Each MR must be grounded in
either \textit{universal logical axioms} or \textit{fundamental physical laws}.
This ensures MRs are model-agnostic and unbiased toward specific architectures.
(2) \textbf{Testability.} Each MR must be verifiable using only spatiotemporal
state data available during embodied execution. This excludes purely cognitive
phenomena that lack observable behavioral correlates. (3) \textbf{Coverage of
Core Spatial Attributes.} The MR set must collectively test the four fundamental
spatial attributes (direction, distance, size, depth) identified in
\textsection{\ref{sec:embodied_spatial_cognition}}.

Based on these criteria, we derived a \textit{minimal sufficient set} of MRs to cover the embodied SC spectrum. 
Our set fully covers the four fundamental SCs identified in \textsection{\ref{sec:embodied_spatial_cognition}} without redundancy to ensure the completeness of our MR set, as each MR targets distinct cognitive mechanisms.
For instance, within directional tasks, MR1 (Transitivity) focuses on multi-hop reasoning (e.g., $s_1 \to s_2 \to s_3$), whereas MR2 (Symmetry) assesses single-step reasoning. Similarly, within magnitude tasks, MR6 (Object Size Ratio Consistency) evaluates relative perception between objects' sizes, whereas MR5 (Size-Depth Consistency) demands adherence to perspective geometry linking projected size to depth. 
Regarding reliability, as demonstrated in \cref{sec:rq1_sc_performance}, the high human baseline (0.96) confirms that our MRs accommodate valid spatial interpretations, ensuring minimal over-flagging. We further provide a quantitative false positive analysis in \cref{sec:false_positive_analysis}.
Finally, our MR design ensures universality. By grounding MRs in fundamental spatial attributes rather than specific tasks, they remain applicable across diverse SC types. For instance, MR1 (Transitivity) applies equally to object relations (SC2-a) and motion perception (SC1-a), ensuring adaptability even as new SC tasks emerge. We empirically validate the effectiveness of these MRs in \cref{sec:ablation_study}.

\begin{figure}[t]
    \centering
    \includegraphics[width=\textwidth]{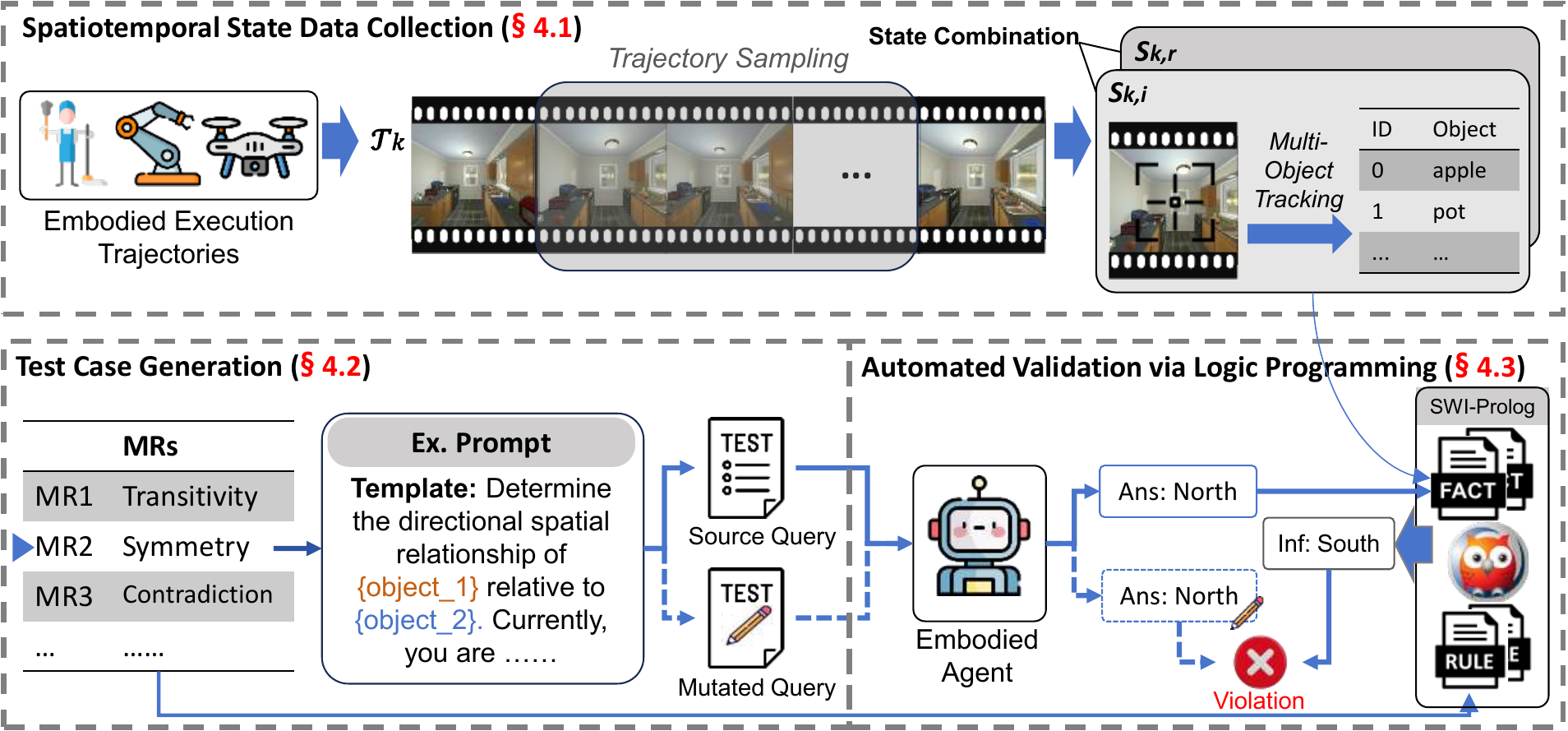}
    \caption{Pipeline of the MetaSpace framework for evaluating embodied spatial cognition.}
    \label{fig:pipeline}
\end{figure}

\section{Methodology}
\label{sec:methodology}
We design and implement MetaSpace to address the challenges mentioned above. The overall architecture of MetaSpace consists of four main modules, as illustrated in \cref{fig:pipeline}:

\begin{itemize}[leftmargin=*, itemsep=1pt, topsep=3pt]
  \item \textbf{Spatiotemporal State Data Collection (\cref{sec:spatiotemporal_state_data_collection}):} Collects and preprocesses real-world trajectory data from embodied agents, sampling state combinations and extracting objects for downstream test case generation.
  \item \textbf{Test Case Generation (\cref{sec:test_case_generation}):} Automatically generates original and transformed test cases from state combinations using metamorphic transformations based on predefined MRs.
  \item \textbf{Automated Validation via Logic Programming (\cref{sec:automated_validation}):} Encodes agent responses and MRs as Prolog facts and rules, and validates consistency between original and transformed outputs to detect violations in embodied spatial cognition responses. 
  \item \textbf{Spatial Capability Scoring and Analysis (\cref{sec:spatial_capability_scoring_and_accuracy_analysis}):} Aggregates violation statistics and computes scores for each spatial cognition capability, enabling detailed evaluation and comparison.
\end{itemize}

\subsection{Spatiotemporal State Data Collection}
\label{sec:spatiotemporal_state_data_collection}
In MetaSpace, we utilize real-world trajectories of embodied intelligence to
evaluate spatial cognition capabilities, ensuring a strong correlation between
evaluation results and actual embodied task performance. Therefore, instead of
relying on static, manually curated test cases, our framework processes real
execution trajectories from embodied agents (more details in
\cref{sec:dataset}). Let \( \mathcal{T} \) represent the set of
raw trajectory data collected from embodied agents performing embodied tasks
(e.g., navigation or manipulation), defined as \( \mathcal{T} = \{
\mathcal{T}_1, \mathcal{T}_2, \ldots, \mathcal{T}_m \} \), where each trajectory
\( \mathcal{T}_k \) is a sequence of states defined as \( \mathcal{T}_k = \{
s_{k,1}, s_{k,2}, \ldots, s_{k,n} \} \). Each state \( s_{k,i} \) includes an
egocentric visual observation by the embodied agent, represented as \( f_{k,i}
\) (image frame). These states are then used to generate test cases (detailed in
\cref{sec:test_case_generation}). We preprocess the trajectory data for
test case generation in the following:

\head{\ding{192}~Trajectory Sampling.} Each trajectory \( \mathcal{T}_k \) is
processed to extract state combinations (e.g., two or three states) based on the
specific SC and MR utilized. Each combination is then used for individual test
case generation. For instance, to adapt \textit{MR2: Symmetry} in \textit{SC1},
we extract pairs of consecutive states \( (s_{k,i}, s_{k,i+1}) \) from the
trajectory \( \mathcal{T}_k \). The implementation details of state combination
extraction are provided in \cref{sec:spatiotemporal_state_data_collection_implementation}.

\head{\ding{193}~Object Detection and Tracking.} For each state combination, we 
extract the objects (e.g., landmarks, manipulable items) for test case
generation. We implement this using Ultralytics YOLO~\cite{redmon2016you} for
multi-object tracking across the whole trajectory. Let \( \mathcal{O}_{k,i} = \{
o_{k,i,1}, o_{k,i,2}, \ldots, o_{k,i,l} \} \) denote the set of objects detected
in a frame \( f_{k,i} \) within trajectory \( \mathcal{T}_k \). Each object \(
o_{k,i,j} \) is detected with its bounding box coordinates, class label, and
confidence score. Since the objective of MetaSpace is to evaluate embodied
spatial cognition capabilities, we aim to minimize interference from perception
capability failures (e.g., the agent failing to recognize the target apple).
Therefore, we adopt the following strategies: 1) Assign a unique identifier (ID)
to each object across frames. 2) Retain only detections with a confidence score
above 0.6 to minimize noise (the threshold determination process is detailed in
\cref{sec:threshold_determination}). 3) Incorporate the detected
bounding boxes into the input prompt for the embodied agent to further reduce
the impact of perception errors. 
We further investigate the performance of YOLO and the robustness against perception noise in \cref{sec:spatiotemporal_state_data_collection_implementation}.

\subsection{Test Case Generation}
\label{sec:test_case_generation}

The state combinations and their corresponding object sets obtained from the \cref{sec:spatiotemporal_state_data_collection} are used to generate test cases. In this module, the trajectories are utilized to generate test cases based on the MRs defined in \cref{sec:logical_MRs} and \cref{sec:physical_MRs}. Each test case consists of an original input query \(x\) and a transformed input query \(x' = \phi(x) \), which is generated by applying the metamorphic transformation \( \phi \) associated with a specific type of MR. Both are derived from the same state combination.

\begin{algorithm}[t]
\caption{Automated Test Case Generation}
\label{alg:test_case_generation}
\begin{algorithmic}[1]
  \small
\Require Trajectory set $\mathcal{T}$, Prompt Templates $\widetilde{template}$, target spatial cognition $SC$
\Ensure Test Cases $\mathcal{C}$
\State $\mathcal{C} \gets [\,]$, $\widetilde{MR} \gets$ \Call{GetMRsBySC}{$SC$}
\For{each trajectory $\mathcal{T}_k$ in $\mathcal{T}$}
    \For{each $MR$ in $\widetilde{MR}$}
        \If{$SC = \text{SC1}$}
            \State $w \gets$ \Call{GetWindowSize}{$MR$}
            \State $\mathcal{S} \gets$ \Call{GenerateConsecutiveCombinations}{$\mathcal{T}_k$, $w$}
        \Else
          \State $\mathcal{P} \gets$ \Call{GetPersistentObjects}{$\mathcal{T}_k$}
          \State $\mathcal{S} \gets$ \Call{GenerateObjectBasedCombinations}{$\mathcal{T}_k, \mathcal{P}$}
        \EndIf
        \For{each state combination $S$ in $\mathcal{S}$}
            \State $filled\_prompt \gets$ \Call{FillTemplate}{$\widetilde{template}[SC,MR]$, $S$, $SC$}
            \State $(x, x') \gets$ \Call{GenerateQueries}{$filled\_prompt$, $MR$}
            \State $\mathcal{C}.\text{append}((x, x', MR))$
        \EndFor
    \EndFor
\EndFor
\State \Return $\mathcal{C}$
\end{algorithmic}
\end{algorithm}

MetaSpace adopts
an automated and spatial cognitive capability-aware approach to sample
trajectories and generate test cases. The detailed process is outlined in
Algorithm~\ref{alg:test_case_generation}. For each trajectory
(\(\mathcal{T}_k\)) in the trajectory set (\(\mathcal{T}\)), and for each MR
associated with the target SC, the algorithm selects state combinations using
different strategies based on the SC type. Specifically, for movement perception
(SC1), a sliding window of size \(w\) (determined by the MR) is applied to
extract consecutive state combinations (Lines 5--6). For SC2, SC3, and SC4, the
algorithm first identifies persistent objects across the trajectory and then
generates object-centric state combinations based on these objects (Lines
8--9). For each state combination, the appropriate prompt template is retrieved
from the predefined set \(\widetilde{template}[SC, MR]\) and filled with the
state combinations and objects (objects are excluded for SC1) (Lines 12).
Subsequently, the original input query \(x\) is generated from the filled prompt
(Line 13). The transformed input query \(x'\) is created by applying the
metamorphic transformation \(\phi\) associated with the current MR to the filled
prompt (Line 13). Finally, the pair \((x, x')\), along with the corresponding
MR, is appended to the test case set \(\mathcal{C}\) (Line 14). After processing
all trajectories and MRs, the algorithm returns the complete set of generated
test cases \(\mathcal{C}\) (Line 18). This approach ensures that test case
generation is fully automated, reproducible, and scalable, while remaining
agnostic to the specific MR or the length of the agent's
trajectories.

\subsection{Automated Validation via Logic Programming}
\label{sec:automated_validation}

MetaSpace automates output validation by encoding spatial observations as Prolog
facts and MRs as rules. For each test case, the Prolog engine infers the
expected output for the transformed input using these facts and rules, then
compares it to the agent's actual response. Any inconsistency indicates a
spatial cognition error. 

As detailed in \cref{alg:consistency_validation}, the process begins with an
automatic rule parser that iterates over all MRs, extracting for each relation a
specific query pattern (a Prolog predicate template for enumerating relevant
test case instances) and the corresponding reasoning rule (line 3). A Prolog
program is constructed by combining the facts from observations with the
reasoning rule for the MR (line 4). Using this program, all possible
instantiations of the MR predicate template, which represent valid combinations
of states or objects present in the ground facts, are enumerated for validation
(line 5). For each instantiation, the algorithm checks consistency between the
expected output inferred by the Prolog engine and the agent's actual response;
if the check fails, a violation is recorded and appended to a set for further
analysis (lines 6--9). This automated process enables MetaSpace to
comprehensively assess the spatial cognitive consistency of the agent across all
possible test cases derived from its observations.
We utilize SWI-Prolog~\cite{wielemaker:2011:tplp}, an open-source advanced logic programming interpreter to implement the Prolog engine.

\begin{algorithm}[t]
\caption{Automated Consistency Validation via Logic Programming}
\label{alg:consistency_validation}
\begin{algorithmic}[1]
  \small
\Require Observations $\widetilde{F}_{obs}$, Metamorphic Relations $\widetilde{MR}$
\Ensure Violation Set $\widetilde{V}$
\State $\widetilde{V} \gets [\,]$ \Comment{Initialization}
\For{each $MR$ in $\widetilde{MR}$} \Comment{Iterate over each MR}
    \State $(Q_{MR}, \mathcal{R}_{MR}) \gets$ \Call{ParseMR}{$MR$} \Comment{Obtain MR-specific query and reasoning rule}
    \State $\mathcal{P} \gets \widetilde{F}_{obs} ++ \mathcal{R}_{MR}$ \Comment{Construct Prolog program with facts and MR rule}
    \State $Inst \gets$ \Call{FindAllInstantiations}{$\mathcal{P}$, $Q_{MR}$} \Comment{Enumerate all entity tuples to check}
    \For{each $inst$ in $Inst$} \Comment{Iterate over each instantiation}
        \If{\textbf{not} \Call{CheckConsistency}{$\mathcal{P}$, $Q_{MR}$, $inst$}}
            \State $V_{new} \gets (MR, inst)$ \Comment{Record violation: MR and instance}
            \State $\widetilde{V}.\text{append}(V_{new})$
        \EndIf
    \EndFor
\EndFor
\State \Return $\widetilde{V}$ \Comment{Return all detected violations}
\end{algorithmic}
\end{algorithm}

\subsection{Spatial Capability Scoring and Accuracy Analysis}
\label{sec:spatial_capability_scoring_and_accuracy_analysis}
After executing all test cases and collecting the violation set \( \widetilde{V}
\), MetaSpace computes a spatial capability score for each spatial cognitive
capability defined in Fig.~\ref{fig:taxonomy}. The scoring process is
defined in the following. For each embodied spatial cognitive capability \( SC_i
\), identify the set of associated metamorphic relations \( \widetilde{MR}(SC_i)
\). Let ${\mathcal{\widetilde{C}}(SC_i)}$ denote the set of all test cases
generated for $SC_i$ based on \( \widetilde{MR}(SC_i) \), and
$\widetilde{V}(SC_i)$ denote the set of violations detected for $SC_i$. Compute
the capability score using the formula:
  \begin{equation}
  Score(SC_i) = 1 - \frac{|\widetilde{V}(SC_i)|}{|\widetilde{\mathcal{C}}(SC_i)|}
  \end{equation}
where $|\widetilde{\mathcal{C}}(SC_i)|$ is the total number of test cases and $|\widetilde{V}(SC_i)|$ is the number of violations. This score ranges from 0 to 1, where a score of 1 indicates perfect performance (no violations), and a score of 0 indicates complete failure (all test cases resulted in violations).

\section{Implementation}
\label{sec:implementation}

\subsection{Dataset}
\label{sec:dataset}
The three embodied scenarios and their corresponding datasets used in our evaluation are as follows:
    \textbf{Household robot:} We use the EB-Navigation dataset from~\cite{yang2025embodiedbench}, which is based on AI2-THOR~\cite{kolve2022ai2thorinteractive3denvironment}. It contains 60 navigation trajectories in different household scenes (e.g., kitchens, living rooms, and bedrooms). 
    \textbf{Robotic arm:} We use the EB-Manipulation dataset from~\cite{yang2025embodiedbench}, which is based on VLMBench~\cite{10.5555/3600270.3600318} using the CoppeliaSim simulator~\cite{6696520} to control a 7-DoF Franka Emika Panda robotic arm. The dataset contains 48 manipulation trajectories in different tabletop scenes.
    \textbf{Drone:} We use a sub-dataset of drone navigation trajectories from~\cite{zhao2025urbanvideo}, which is based on the AerialVLN~\cite{10378183}. The sub-dataset contains 50 navigation trajectories in various outdoor scenes (e.g., urban areas, parks, and highways).

\subsection{Benchmark Agents}
\label{sec:benchmark_models}
\head{Model Selection.}
To ensure reliable evaluation, we assess embodied agents powered by 6 SOTA MLLMs using MetaSpace. We select two categories for analysis: (1) closed-source, API-accessible models, including GPT-5~\cite{openai2025gpt5}, GPT-4o~\cite{openai2024gpt4o}, and Claude Sonnet 4~\cite{anthropic2025claude}; and (2) open-source, locally deployable models, including Qwen-VL~\cite{bai2023qwenvlversatilevisionlanguagemodel}, InternVL3.5-8B~\cite{wang2025internvl35advancingopensourcemultimodal}, and DeepSeek-VL2-small~\cite{wu2024deepseekvl2mixtureofexpertsvisionlanguagemodels}.

\head{Model Configurations.}
To ensure the stability and consistency of model outputs during evaluation, we set the \textit{temperature} parameter to 0, resulting in deterministic responses. We further set \textit{top-p} to 0.9 and disable \textit{top-k} sampling (set to 0), so that only the most probable tokens are selected, thereby improving the reliability of generated results. 

\head{Consistency in Model Outputs.}
To rigorously assess the consistency of MLLM responses in our approach, we conduct statistical significance tests. Specifically, for each embodied scenario and SC, we randomly sample 30 test cases, yielding a total of 720 test cases. Taking GPT-4o as an example, each test case was executed five times under the above configuration. To evaluate whether the responses from different runs were statistically indistinguishable, we then applied the Friedman test~\cite{Friedman01121937}, a non-parametric method for detecting differences across multiple repeated measures. The results showed no significant differences between runs (average $p$-value = 0.57), confirming that the MLLM outputs are highly consistent under our settings. Therefore, a single run is sufficient for evaluation in MetaSpace, ensuring both efficiency and reliability.

\begin{table*}[t]
    \centering
    \begin{minipage}{0.48\textwidth}
        \centering
        \caption{MR5 Threshold Sensitivity Analysis (Score).}
        \label{tab:mr5_sensitivity}
        \resizebox{\linewidth}{!}{% 自适应宽度
            \begin{tabular}{lcccc}
                \toprule
                \textbf{Threshold Variation} & $\epsilon$ & \textbf{Human} & \textbf{GPT-4o} & \textbf{Gap} \\
                \midrule
                Strict (-50\%)   & 0.050 & 0.88 & 0.39 & 0.49 \\
                Strict (-25\%)   & 0.075 & 0.91 & 0.41 & 0.50 \\
                \textbf{Original (Ref)} & \textbf{0.100} & \textbf{0.93} & \textbf{0.44} & \textbf{0.49} \\
                Relaxed (+25\%)  & 0.125 & 0.95 & 0.45 & 0.50 \\
                Relaxed (+50\%)  & 0.150 & 0.97 & 0.48 & 0.49 \\
                \bottomrule
            \end{tabular}%
        }
    \end{minipage}
    \hfill % 在两个表格中间填充空格，把它们推向两边
    \begin{minipage}{0.48\textwidth}
        \centering
        \caption{MR6 Threshold Sensitivity Analysis (Score).}
        \label{tab:mr6_sensitivity}
        \resizebox{\linewidth}{!}{% 自适应宽度
            \begin{tabular}{lcccc}
                \toprule
                \textbf{Threshold Variation} & $\delta$ & \textbf{Human} & \textbf{GPT-4o} & \textbf{Gap} \\
                \midrule
                Strict (-50\%)   & 0.0250 & 0.86 & 0.75 & 0.11 \\
                Strict (-25\%)   & 0.0375 & 0.88 & 0.76 & 0.12 \\
                \textbf{Original (Ref)} & \textbf{0.0500} & \textbf{0.91} & \textbf{0.79} & \textbf{0.12} \\
                Relaxed (+25\%)  & 0.0625 & 0.93 & 0.82 & 0.11 \\
                Relaxed (+50\%)  & 0.0750 & 0.94 & 0.84 & 0.10 \\
                \bottomrule
            \end{tabular}%
        }
    \end{minipage}
\end{table*}

\subsection{Spatiotemporal State Data Collection}
\label{sec:spatiotemporal_state_data_collection_implementation}
\head{Trajectory Sampling Strategy.}
The trajectory sampling strategy varies based on the SC capability being evaluated. For movement perception evaluation, we exclusively utilize consecutive states to assess agents' immediate motion perception rather than their long-term path planning abilities. This distinction enables embodied agents to support step-by-step action decisions through continuous perception-action loops, while higher-level planning capabilities rely on goal-oriented strategies for long-term navigation. This approach emphasizes our focus on core SC skills rather than the extensively discussed path planning capabilities~\cite{kong2024embodiedaimobilerobots,zhang2024met}.
For evaluations of spatial reasoning, perspective visualization, and egocentric-allocentric transformation, we do not restrict to consecutive states, as these capabilities have different requirements. Spatial reasoning requires understanding the relationships between objects in space, often necessitating multiple time steps to observe the same objects from various angles. Perspective visualization often requires observations from multiple viewpoints that may not be captured in consecutive frames. Additionally, egocentric-allocentric transformation requires agents to integrate information from different spatiotemporal positions to construct coherent mental maps. Therefore, we extract different state combinations from the trajectories based on the SC being evaluated.

\head{Object Detection and Tracking.}
We utilize YOLOv11~\cite{yolo11_ultralytics} for multi-object tracking in the collected spatiotemporal states.
Notably, MetaSpace utilizes YOLO solely for candidate object discovery to generate queries, rather than defining the ground-truth of spatial relations. The oracle of MetaSpace relies on internal logical consistency (e.g., Transitivity) or relative physical ratios, independent of absolute YOLO coordinates. Therefore, potential missed objects (due to recall limitations) merely reduce the quantity of generated test cases without introducing false violations.
We evaluated YOLO's tracking performance on a dataset of 500 images across 50 episodes. The results demonstrate that YOLOv11 achieves 94\% IoU without false positives, ensuring minimal perception errors.
To further validate robustness against perception noise, we injected synthetic noise (including label inaccuracies and bounding box jitter of up to $\pm 20\%$) into 1,000 test instances. 
We observed only six cognitive failures resulting from this noise, which were attributed to the MLLM's inherent perception capabilities for self-correction.
This confirms that perception noise has a negligible impact on our cognitive evaluation, and our modular design allows MetaSpace to adopt future perception advancements. We discuss the threshold for YOLO in \cref{sec:threshold_determination}.

\subsection{Test Case Generation Implementation}
\label{sec:test_case_generation_impl}
In total, we generate 30,300 unique test cases. This number is derived by exhaustively enumerating all valid state and object combinations across all scenarios, SC capabilities, and MRs, as determined by the automated test case generation process in \cref{sec:test_case_generation}. 
These test cases comprehensively cover the eight SCs outlined in \cref{fig:taxonomy}. For each SC, we create test cases based on the associated MRs defined in \cref{sec:logical_MRs} and \cref{sec:physical_MRs}. We then utilize these test cases to evaluate the six benchmark embodied agents.

\subsection{Threshold Determination}
\label{sec:threshold_determination}
To determine the optimal threshold for MR5 and MR6, the criterion is to ensure that the embodied agent achieves human-level performance.
To establish a rigorous baseline, we recruited five adult participants, and each participant was presented with the same set of SC tasks and visual stimuli as the evaluated agents, under standardized instructions and conditions. Participants completed the tasks independently, without time limits or external assistance, to minimize potential biases. All responses were collected and evaluated using the same automated logic-based validation framework applied to the agents. 
The five participants exhibited high inter-subject consistency with low standard deviations (approx. $0.05$ for MR5 and approx. $0.025$ for MR6), indicating that increasing the participant count would unlikely shift the derived thresholds significantly.
By analyzing the relative errors between the participants' estimates and the ground truth, the selected thresholds were calibrated to approximately the 90th percentile of human precision. This allows for minor estimation deviations while rigorously capturing SC capabilities. Consequently, we set the relative error threshold for MR5 ($\epsilon$) and MR6 ($\delta$) to 0.1 and 0.05, respectively. This implies that we accept a 10\% relative error for MR5 and 5\% for MR6 as the bounds of human-level performance; any estimates exceeding these thresholds are considered violations.
We further quantified robustness by testing human and GPT-4o performance under threshold variations ($\pm 25\%$, $\pm 50\%$). As shown in Tab.~\ref{tab:mr5_sensitivity} and Tab.~\ref{tab:mr6_sensitivity}, the GPT-4o SC score improved by only approximately 0.05 even when relaxing the threshold by 50\%. 
This suggests that the detected errors are catastrophic logic violations rather than marginal misses; therefore, slight changes in thresholds do not alter our research conclusions.
To determine the optimal confidence score threshold for multi-object tracking using YOLO, we conduct experiments on 500 images from 50 episodes, testing thresholds ranging from 0.4 to 0.9. Manual inspection of the detection results indicates that a threshold of 0.6 yields an average of four detected objects per image, with a precision of 100\% and a recall of 82\%. This threshold ensures both low false positive rates and adequate object coverage for spatial cognition evaluation.

\section{Evaluation}
\label{sec:experiments}
Our evaluation aims to answer the following research questions (RQs):

\begin{itemize}[leftmargin=*, itemsep=0pt, topsep=0pt]
    \item \textbf{RQ1 (SC Performance): How do different MLLM-driven embodied agents perform in embodied spatial cognition?} This question evaluates the embodied spatial cognition capabilities of various MLLM-driven embodied agents by using MetaSpace.
    
    \item \textbf{RQ2 (Comparison with Existing Works): How does MetaSpace compare with existing approaches in evaluating embodied SC capabilities?} This RQ studies whether MetaSpace outperforms existing benchmarks from both quantitative and qualitative perspectives.
    
    \item \textbf{RQ3 (Internal Evaluation): How effective are individual MRs and how reliable is the overall detection mechanism?} This RQ investigates the effectiveness of each MR in detecting embodied spatial cognition errors and assesses the reliability of the entire detection framework.

    \item \textbf{RQ4 (Mitigation): How can we mitigate the limitations of current MLLM-driven embodied agents in spatial cognition tasks?} This RQ explores potential strategies for improving the performance of embodied agents in spatial cognition tasks.
\end{itemize}

\subsection{Experimental Setup}
\label{sec:experimental_setup}
Our experiments are conducted on a server running Ubuntu 22.04, equipped with
dual 56-core Intel Xeon Scalable processors, \SI{2}{TB} of RAM, and an NVIDIA H800 GPU
node. The total GPU hours consumed for all experiments on open-source MLLMs (all scenarios) amount to \SI{31660.412}{seconds}, which is acceptable given the scale of our evaluation.

\begin{figure}[t]
    \centering
    \begin{minipage}{0.43\textwidth}
        \centering
        \includegraphics[width=\linewidth]{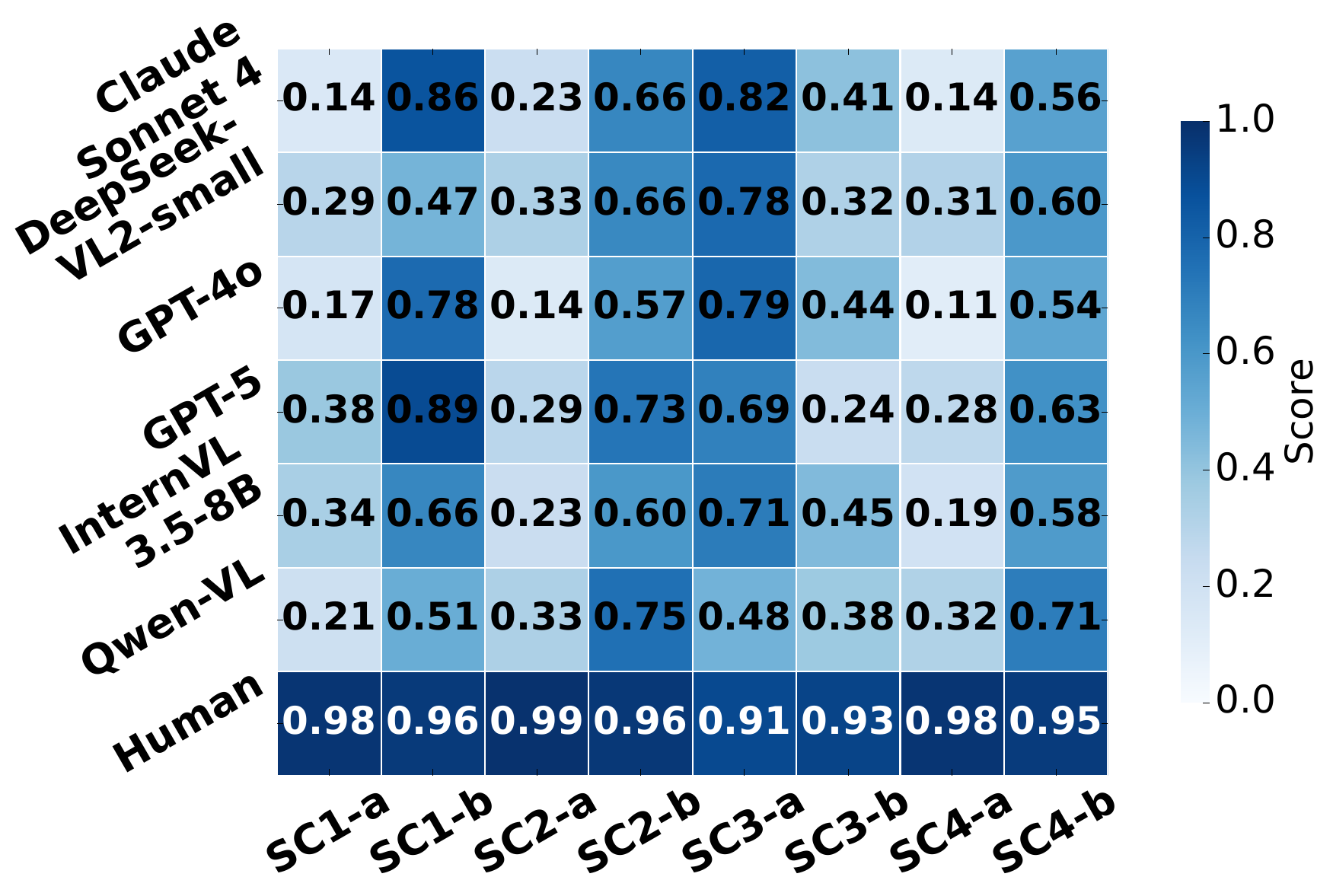}
        \caption{Heatmap of embodied spatial cognition scores across different embodied agents.}
        \label{fig:heatmap}
    \end{minipage}
    \hfill
    \begin{minipage}{0.51\textwidth}
        \centering
        \includegraphics[width=\linewidth]{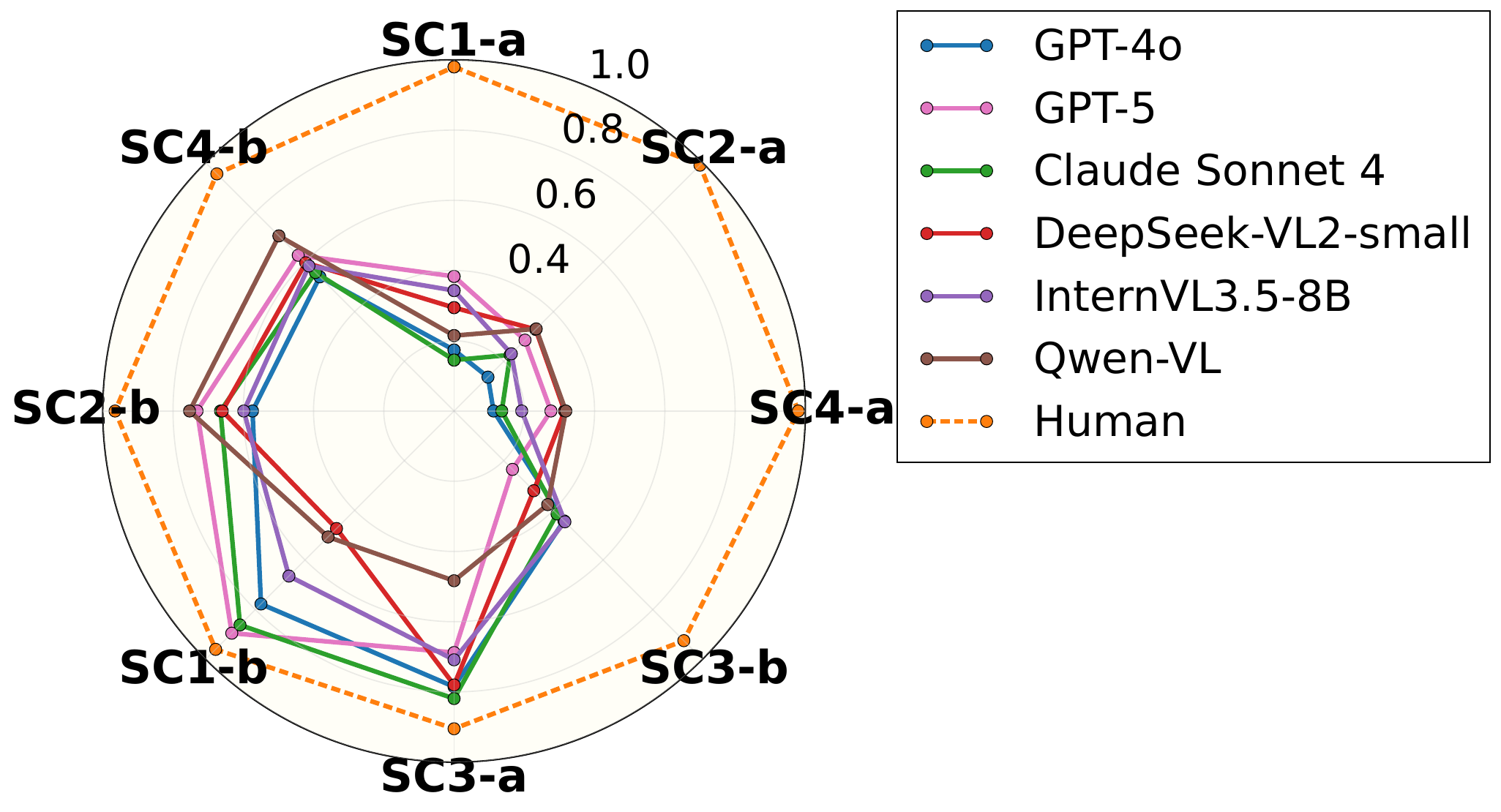}
        \caption{Radar chart comparing embodied spatial cognition scores among different embodied agents.}
        \label{fig:radar}
    \end{minipage}
\end{figure}

\subsection{RQ1: SC Performance}
\label{sec:rq1_sc_performance}
To evaluate the embodied spatial cognition capabilities of six benchmark MLLM-driven embodied agents, we analyze the statistics of test cases and violations detected by MetaSpace. The detailed embodied spatial cognition scores for each agent across all SCs are displayed in Fig.\ref{fig:heatmap} and Fig.~\ref{fig:radar}.
Additionally, Tab. \ref{tab:normalized_stats} reports normalized error statistics. The high failure density (e.g., $>$82 errors per trajectory) verifies that the substantial volume of detected errors stems from pervasive cognitive failures throughout task execution, rather than being an artifact of the test case scale.

\head{Magnitude vs. Directional Tasks.}
Results show that direction estimation and directional spatial reasoning tasks (e.g., movement direction estimation in SC1-a, directional reasoning in SC2-a and SC4-a) are relatively challenging for most agents compared to magnitude estimation tasks (e.g., distance estimation in SC1-b, SC2-b, SC4-b, size estimation in SC3-a, and depth estimation in SC3-b). This phenomenon can be attributed to the fact that magnitude-related tasks primarily require direct numerical inference and quantitative reasoning based on visual input, without necessitating complex spatial relationship modeling. Current mainstream MLLMs are pre-trained on large-scale corpora that emphasize static visual descriptions, object attribute recognition, and fundamental quantitative judgments, making these capabilities more readily generalizable to magnitude estimation tasks~\cite{chatterjee2024gettingrightimprovingspatial,zhu2023multimodalc4openbillionscale}.
In contrast, direction estimation and directional spatial reasoning tasks require advanced spatial representations and reasoning chains in three-dimensional space, which are not yet sufficiently developed in existing MLLMs~\cite{chatterjee2024gettingrightimprovingspatial,hoehing2023whatsleftcantright,ramakrishnan2025doesspatialcognitionemerge,zhang2025scalingbeyondadvancingspatial}. As a result, while agents perform relatively well on magnitude tasks, their poor performance on directional SC highlights the limitations in spatial reasoning capabilities of current MLLM-driven embodied agents.

\head{Size Estimation (SC3-a).} Among all spatial cognition tasks, most agents demonstrate outstanding performance in SC3-a (size estimation), with all models except Qwen-VL scoring above 0.7. This can be attributed to the robust spatial common sense knowledge base of MLLMs, which includes typical dimensions of common objects. By referencing nearby objects with known sizes, MLLMs can effectively estimate the sizes of unknown objects. This strategy renders size estimation tasks relatively easier for MLLM-driven agents.

\head{Egocentric-Allocentric Transformation (SC4).} SC4 evaluates the agent's ability to construct mental maps and perform reference frame transformations. Similar to SC2 (spatial reasoning), SC4 involves reasoning about spatial relationships, but places greater emphasis on transforming from a first-person to an allocentric perspective. This requires the agent not only to comprehend spatial relationships between objects, but also to map these relationships across different viewpoints. Given the complexity involved, all MLLM-driven agents perform poorly on this capability, with scores consistently lower than those achieved on SC2 tasks. This highlights the current limitations of MLLMs in handling dynamic reference frame changes and constructing mental maps.

\begin{figure}[t]
    \centering
    \begin{minipage}{0.48\textwidth}
        \centering
        \includegraphics[width=\linewidth]{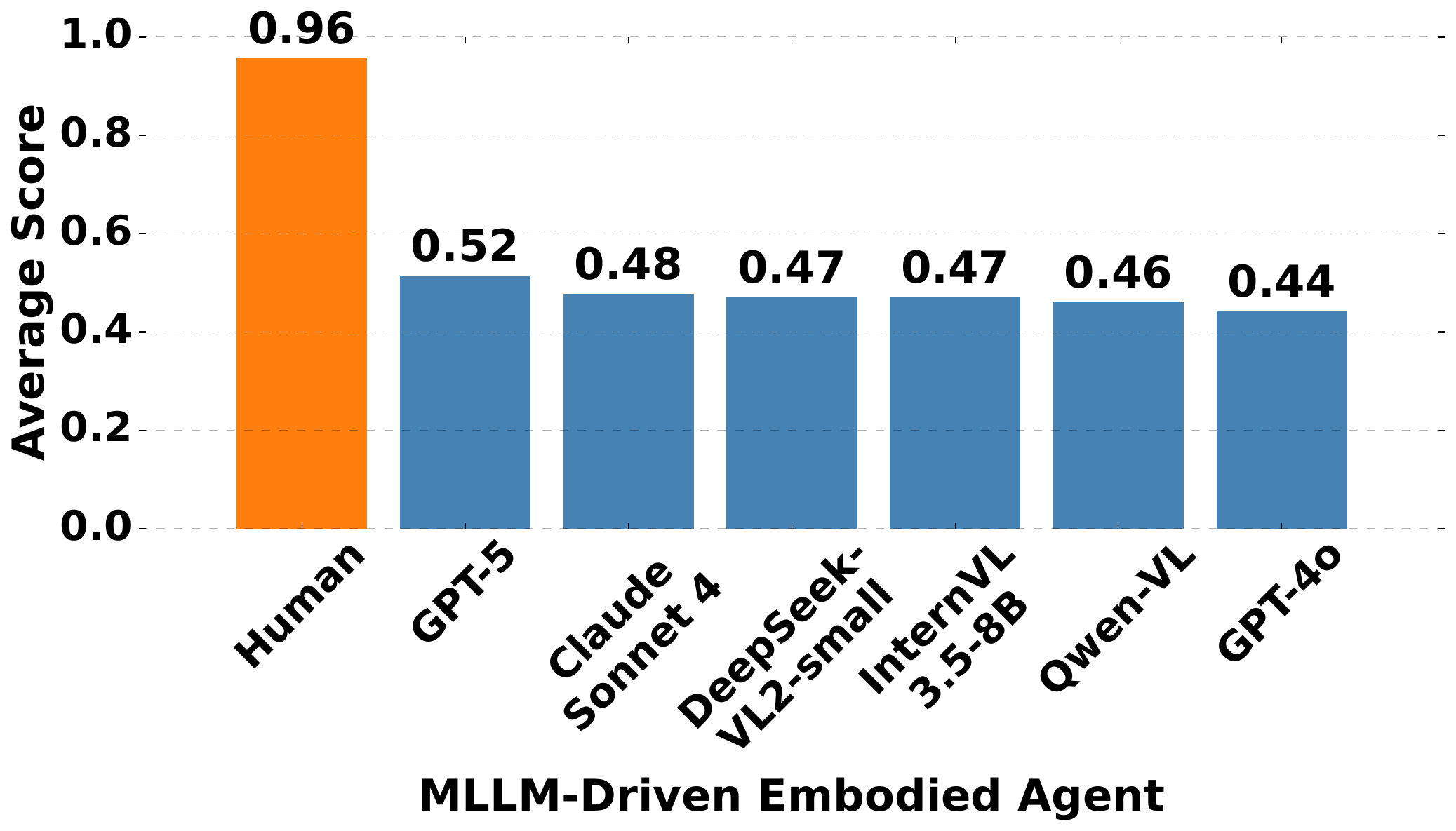}
        \caption{Ranking of average scores for embodied spatial cognition capabilities across various agents.}
        \label{fig:ranking}
    \end{minipage}
    \hfill
    \begin{minipage}{0.48\textwidth}
        \centering
        \includegraphics[width=\linewidth]{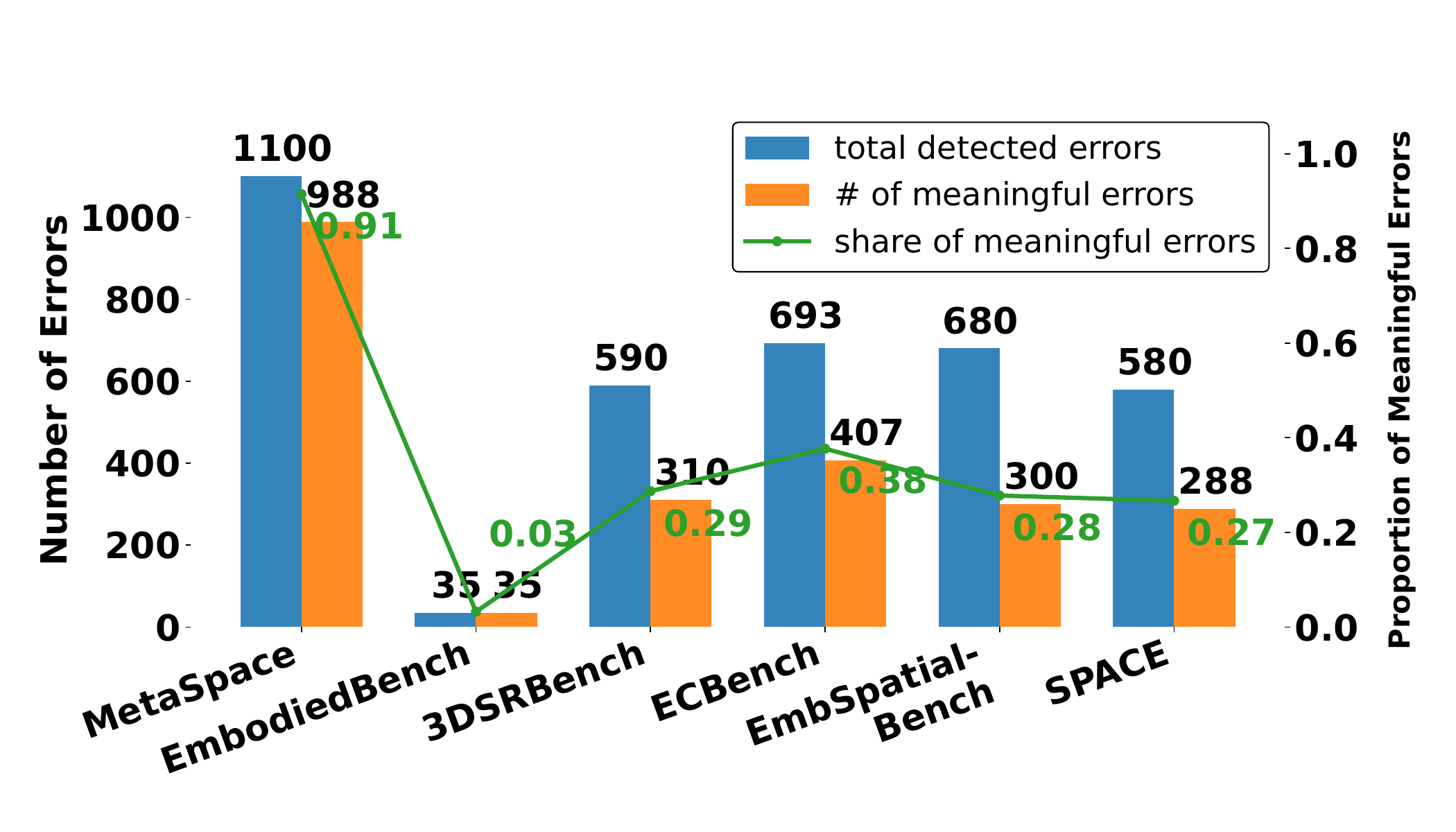}
        \caption{Quantitative comparison of MetaSpace with existing approaches in SC error detection.}
        \label{fig:comparison_with_existing_works}
    \end{minipage}
\end{figure}

\head{Human Baseline Performance.} To further contextualize the overall performance of embodied agents in spatial cognition, we invite five human participants (none of whom have any known spatial cognition impairments) to complete the same set of test cases, serving as a baseline for comparison.\footnote{Considering the large scale of the test cases, which would create a significant workload for human respondents, we decided to conduct the evaluation on a subset comprising 10\% of the original dataset.} As shown in Fig.~\ref{fig:heatmap} and Fig.~\ref{fig:radar}, humans consistently outperform all embodied agents across all spatial cognition tasks, underscoring the significant gap between current MLLM-driven embodied agents and human-level spatial cognition.

\head{Ranking of Average Scores.}
In Fig.~\ref{fig:ranking}, we present the ranking of average scores for embodied spatial cognition capabilities across different embodied agents. The agent powered by the proprietary model GPT-5 achieves the highest average score of 0.52, indicating relatively robust spatial cognition abilities. Notably, open-source models (e.g., DeepSeek-VL2-small and InternVL3.5-8B) also demonstrate competitive performance, with an average score of 0.47, surpassing the proprietary model GPT-4o. While there are variations in average scores among MLLM-driven agents, these discrepancies are relatively minor, with all models scoring between 0.44 and 0.52. In contrast, the human baseline achieves an average score of 0.96, highlighting a substantial gap between embodied agents and human performance.
We further clarify that similar rankings in Fig.~\ref{fig:ranking} result from averaging scores across all SCs. This reflects the generally limited cognitive capabilities of current agents and highlights the substantial human-agent performance gap. However, granular diagnostics for each agent are still evident in Fig.~\ref{fig:heatmap}, such as identifying GPT-5's specific weakness in SC3-b. By leveraging MT with fine-grained MRs to comprehensively assess four SCs, MetaSpace mitigates the ``success by coincidence'' issue found in existing benchmarks and provides granular diagnostics that binary success metrics in other benchmarks cannot reveal.

\begin{table*}[t]
  \centering
  \caption{
    Normalized error statistics. \textbf{Err. Rate}: Error Rate (\%, per test case); \textbf{Avg. Err. Tra.}: Average Errors per Trajectory; \textbf{Avg. Err. Obj.}: Average Errors per Object.
  }
  \label{tab:normalized_stats}
  \small
\begin{tabular}{lcccccc}
    \toprule
    & \multirow{2}{*}{\textbf{GPT-5}} & \textbf{Claude} & \multirow{2}{*}{\textbf{Qwen-VL}} & \textbf{InternVL} & \textbf{DeepSeek} & \multirow{2}{*}{\textbf{GPT-4o}} \\ 
    & & \textbf{Sonnet 4} & & \textbf{3.5-8B} & \textbf{VL2-small} & \\
    \midrule
    \textbf{Total Test Cases} & \multicolumn{6}{c}{30,300 per agent} \\
    \midrule
    \textbf{Total Errors} & 13,048 & 14,527 & 15,024 & 15,636 & 15,737 & 16,449 \\ 
    \textbf{Err. Rate} & 43.06\% & 47.94\% & 49.58\% & 51.60\% & 51.94\% & 54.29\% \\
    \textbf{Avg. Err. Tra.} & 82.58 & 91.94 & 95.09 & 98.96 & 99.60 & 104.11 \\
    \textbf{Avg. Err. Obj.} & 16.52 & 18.39 & 19.02 & 19.79 & 19.92 & 20.82 \\
    \bottomrule
  \end{tabular}
\end{table*}

\begin{tcolorbox}[title=ANSWER to RQ1, boxrule=0.8pt,boxsep=1.5pt,left=2pt,right=2pt,top=2pt,bottom=1pt]
Our evaluation using MetaSpace reveals that benchmark MLLM-driven embodied agents exhibit significant limitations in embodied spatial cognition, particularly in tasks involving direction estimation and spatial directional reasoning. While these agents perform relatively well on magnitude estimation tasks, their overall scores remain substantially lower than human performance.
\end{tcolorbox} 

\subsection{RQ2: Comparison with Existing Works}

\subsubsection{Qualitative Analysis}
We qualitatively compare MetaSpace with the SOTA embodied agents benchmarks and embodied spatial cognition evaluation approaches to illustrate the advantages of MetaSpace. As shown in Tab.~\ref{tab:comparison_with_existing_works}, we compare MetaSpace with EmbodiedBench~\cite{yang2025embodiedbench}, 3DSRBench~\cite{ma20253dsrbenchcomprehensive3dspatial}, ECBench~\cite{Dang_2025_CVPR}, EmbSpatial-Bench~\cite{du-etal-2024-embspatial}, and SPACE~\cite{ramakrishnan2025doesspatialcognitionemerge} from four core dimensions.

\head{Test Case Generation.} Existing benchmarks primarily rely on manually annotated VQA or MCQ datasets, which are labor-intensive and costly to create, limiting their scalability. Besides, the quality of manually created test cases can vary significantly due to human subjectivity, leading to potential biases and inconsistencies. In contrast, MetaSpace employs an automated test case generation approach based on MRs, enabling the creation of a vast number of test cases without human intervention. This automation not only enhances scalability but also ensures reproducibility and consistency in test case generation. 

\head{Embodiment.} Some existing benchmarks (e.g., 3DSRBench, EmbSpatial-Bench) focus on static VQA tasks without considering the embodied nature of agents, which limits their ability to evaluate spatial cognition in embodied settings. For example, SPACE includes some classic human cognitive tests (e.g., Minnesota Paper Form Board test), which do not involve any interaction with the environment. In contrast, MetaSpace is specifically designed for testing in real embodied settings, by utilizing the real embodied execution trajectories of agents to generate test cases. Our approach allows for a more realistic embodiment evaluation of spatial cognition capabilities.

\head{Test Oracle.} Existing benchmarks typically use high-level task success, VQA/MCQ correctness (by comparing with human-annotated answers), or human evaluation as the test oracle. However, high-level task success can miss errors that do not directly cause task failure (Fig.~\ref{fig:false_positive_success}), while manual annotation and human evaluation are labor-intensive and may introduce noise. In contrast, MetaSpace uses violations of MRs as the test oracle. These MRs are grounded in established logic rules and physical laws, providing an objective, scalable, and reliable validation process.

\head{Spatial Cognition Categories.} Existing benchmarks typically emphasize SC2 (spatial reasoning), with limited coverage of other essential SC capabilities (e.g., SC1, SC3, and SC4). This narrow focus restricts the ability to holistically evaluate embodied agents. In contrast, MetaSpace provides comprehensive coverage of all eight key SC capabilities, enabling a more complete and robust assessment of embodied SC. We will further discuss the necessity of these SCs in \textsection{\ref{sec:quantitative_analysis}}.

\begin{table}[t]
    \caption{Qualitative comparison of MetaSpace with existing SOTA evaluation approaches.}
    \label{tab:comparison_with_existing_works}
    \centering
    \small
    \begin{tabular}{c c c c c}
        \toprule
        & \textbf{Test Case Generation} & \textbf{Embodiment} & \textbf{Test Oracle} & \textbf{SC Categories} \\ 
        \midrule
        MetaSpace & Auto-gen based on MRs & \Checkmark & MR violation & SC1, SC2, SC3, SC4 \\ 
        \midrule
        EmbodiedBench & \centering - & \Checkmark & Task success & SC2, Perception \\ 
        \midrule
        3DSRBench & Man. annotated VQA &  & VQA correctness & SC2, \footnotesize Height estimation \\ 
        \midrule
        \footnotesize EmbSpatial-Bench & \begin{tabular}[c]{@{}l@{}}Auto-gen MCQs\end{tabular} &  & MCQ correctness & SC2-a, SC3-b \\ 
        \midrule
        \multirow{2}{*}{ECBench} & \multirow{2}{*}{Man. annotated VQA} & \multirow{2}{*}{\Checkmark} & \begin{tabular}[c]{@{}l@{}}VQA correctness,\end{tabular} & \begin{tabular}[c]{@{}l@{}}SC2, Perception,\end{tabular} \\ 
         & & & Human eval. & Hallucination \\
        \midrule
        \multirow{2}{*}{SPACE} & \begin{tabular}[c]{@{}l@{}}Man. annotated MCQs,\end{tabular} & \multirow{2}{*}{$\sqrt{}\mkern-9mu{\smallsetminus}$} & \begin{tabular}[c]{@{}l@{}}MCQ correctness,\end{tabular} & \begin{tabular}[c]{@{}l@{}} \footnotesize Classic cognitive tests, \end{tabular} \\ 
         & Interactive tasks &  & SPL & SC2  \\
        \bottomrule
    \end{tabular}
\end{table}

\subsubsection{Quantitative Analysis}
\label{sec:quantitative_analysis}
To compare MetaSpace with existing SOTA approaches in testing SC capabilities of embodied agents, we do a small-scale quantitative analysis. The challenge is that different methods may target distinct aspects of spatial cognition, making direct quantitative comparison among detection results impractical and unfair. Therefore, we design a heuristic evaluation protocol to solve this challenge. Specifically, we apply MetaSpace and other approaches to the trajectories executed by the same embodied agent (these trajectories are not utilized in our framework to avoid data leakage), and collect the union of all detected errors. Next, we invite five independent human reviewers, who have no involvement in the design of our framework and possess extensive experience in embodied intelligence, to annotate whether each detected error is \emph{meaningful} for embodied tasks. The annotation criteria focus on whether the error could negatively impact embodied task performance. These negative impacts include: (1) errors that directly cause task failure, (2) errors that reduce task efficiency, (3) errors that may lead to safety risks in real-world deployment, and (4) errors that did not cause failure in the current task due to coincidence but represent potential risks in similar cases. This protocol allows us to evaluate the proportion of truly \emph{meaningful} spatial cognition errors detected by each method fairly. Finally, we calculate the ratio of \emph{meaningful} errors detected by each method and show results in \cref{fig:comparison_with_existing_works}.

From the results, the union of \emph{meaningful} errors detected by all methods has a total of 1080 errors. Among them, MetaSpace detects 988 \emph{meaningful} errors, accounting for 91.5\% of the union. 
This corresponds to a false negative rate of approximately 8.5\%, substantially lower than other SOTA approaches.
The remaining 8.5\% stems from scenarios where the agent's spatial reasoning is \textit{consistent with MRs yet factually incorrect} (e.g., claiming A is left of B and B is right of A while ground truth is A is right of B). Such errors are only detectable by human-annotated strong oracles. 
In contrast, 3DSRBench detects 310 \emph{meaningful} errors (29\%), ECBench detects 407 \emph{meaningful} errors (38\%), EmbSpatial-Bench detects 300 \emph{meaningful} errors (28\%), and SPACE detects 288 \emph{meaningful} errors (27\%). It is notable that EmbodiedBench only detects 35 spatial cognition errors, of which all are \emph{meaningful}, but the total number is very small. This is because EmbodiedBench primarily focuses on high-level task success as the test oracle, which overlooks many spatial cognition errors that do not directly lead to task failure. 
Crucially, since all approaches were evaluated on identical trajectories, the resulting error concentration in MetaSpace is not due to experimental bias.
Instead, it reflects MetaSpace's capability to test a broader spectrum of SCs than other SOTAs.
These results demonstrate that MetaSpace significantly outperforms existing approaches in detecting spatial cognition errors that are truly \emph{meaningful} for embodied task performance. We provide a more detailed comparison with related testing efforts in \cref{sec:related_work}.

\begin{tcolorbox}[title=ANSWER to RQ2, boxrule=0.8pt,boxsep=1.5pt,left=2pt,right=2pt,top=2pt,bottom=1pt]
Compared to existing SC evaluation approaches, MetaSpace offers significant advantages in test case generation, embodiment, and test oracle design. Quantitatively, MetaSpace detects a substantially higher proportion of spatial cognition errors that impact embodied task performance.
\end{tcolorbox} 

\subsection{RQ3: Internal Evaluation}
\label{sec:internal_evaluation}
To investigate the effectiveness of individual MRs in detecting embodied SC errors and to assess the reliability of MetaSpace, we conduct an internal evaluation.
Firstly, to understand the effectiveness of different MRs, we conduct an ablation study in \cref{sec:ablation_study}. After that, in \cref{sec:false_positive_analysis}, we perform a false positive analysis to assess the accuracy of MetaSpace in detecting embodied SC errors.

\begin{figure}[t]
    \centering
    \begin{minipage}{0.48\textwidth}
        \centering
        \includegraphics[width=\linewidth]{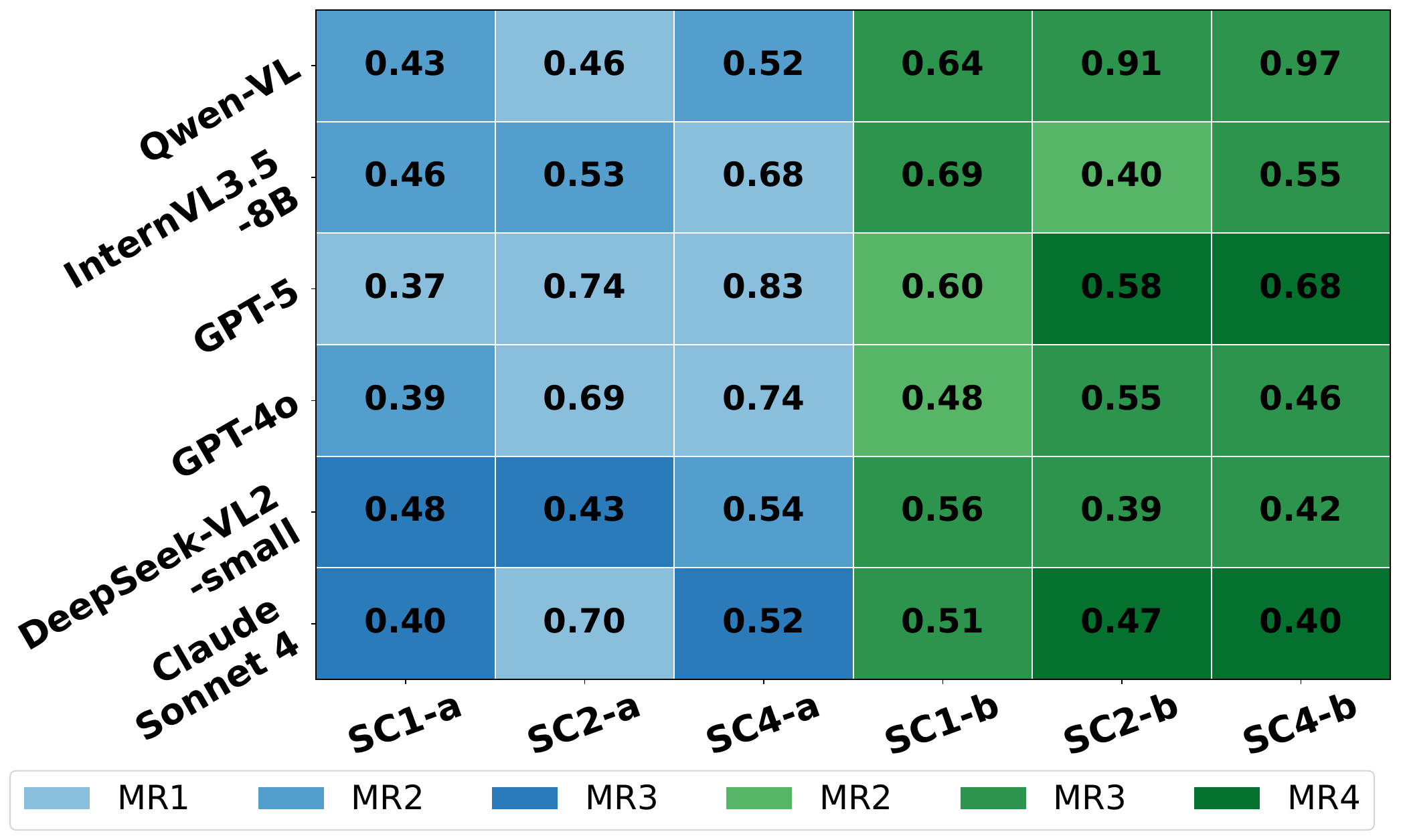}
        \caption{MRs that trigger the most SC errors on diverse agents across SCs. The number in each cell represents the triggered SC error ratio of the corresponding MR type, calculated by $|\text{Errors}(MR_u, SC_v)| \, / \, |\text{Total Errors}(SC_v)|$.}
        \label{fig:rq3-b}
    \end{minipage}
    \hfill
    \begin{minipage}{0.48\textwidth}
        \centering
        \includegraphics[width=\textwidth]{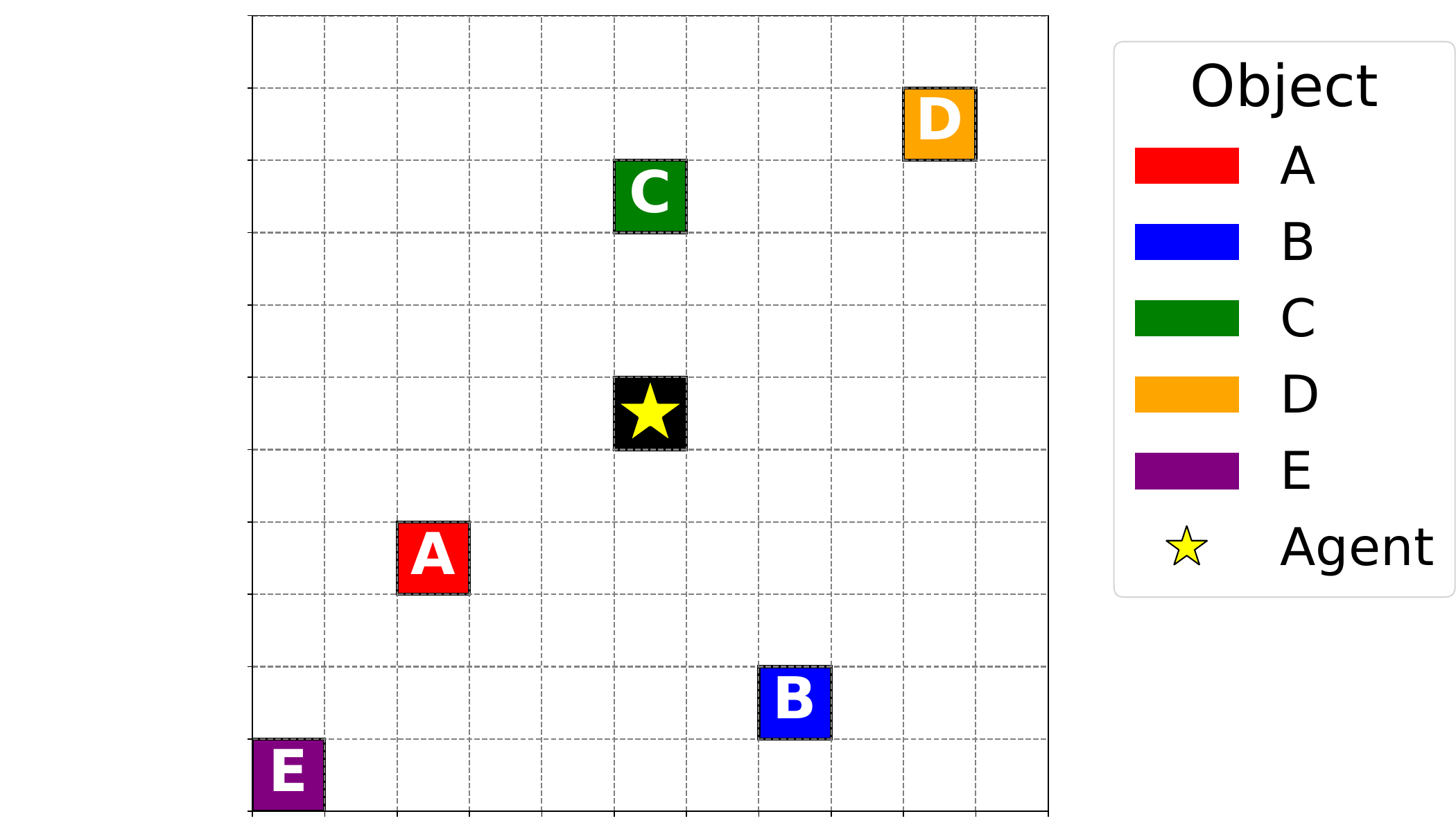}
        \caption{An example of a cognitive map, where the agent predicts object/landmark positions within an 11 $\times$ \SI{11}{grid}, with the agent located at the center of the map. }
        \label{fig:cognitive_map}
    \end{minipage}
\end{figure}

\subsubsection{Ablation Study}
\label{sec:ablation_study}
To assess the effectiveness of different MRs in detecting embodied spatial cognition errors, we conduct an ablation study. For better visualization and understanding, we present the distribution of embodied spatial cognition errors discovered with various MRs in Fig.~\ref{fig:rq3-b}. This figure illustrates which MR is able to identify more spatial cognition errors for different SCs and agents.
The blue section represents the error detection ratio of \textit{MR1}, \textit{MR2} and \textit{MR3} in directional SCs (i.e., SC1-a, SC2-a, SC4-a), while the green section shows the number of errors detected by \textit{MR2}, \textit{MR3}, and \textit{MR4} in magnitude SCs (i.e., SC1-b, SC2-b, SC4-b). \textit{MR1} identifies relatively more spatial cognition errors in directional tasks, whereas MR3 detects a relatively higher number of errors in magnitude SCs. Overall, all types of MR identify a significant number of spatial cognition errors across various embodied spatial cognition types, so we can conclude that all MRs contribute meaningfully to the overall evaluation of embodied spatial cognition.

\subsubsection{False Positive Analysis}
\label{sec:false_positive_analysis}
To evaluate the accuracy of MetaSpace in detecting embodied SC errors, we conducted a manual validation of a randomly selected subset of 1,000 detected errors from the total set of violations \( \widetilde{V} \). The results indicate that MetaSpace produced nine false positives, achieving a precision rate of 99.1\%. 
Adjusting for this 0.9\% FP rate shifts overall scores negligibly (approx. $0.005$), leaving our core conclusions unaltered. Similarly, other SOTAs mentioned in \cref{sec:quantitative_analysis} also suffer from FPs due to annotator inconsistency and subjective bias.

Notably, all false positive instances were concentrated in the evaluations of MR5 and MR6.
In these two MRs, we employed fixed thresholds (e.g., \( \epsilon = 0.1 \) for MR5 and \( \delta = 0.05 \) for MR6) to quantify violations, which were derived from human baselines. However, it is important to recognize that in extreme test cases, human participants often struggle to make precise judgments. This can lead to larger estimation discrepancies and instances where thresholds exceed \( \epsilon = 0.1 \). 
While this limitation is acknowledged, it is also a necessary trade-off to achieve the benefits of quantitative, automated verification. Overall, the results underscore MetaSpace's high precision in advancing error detection in embodied spatial cognition.
We discuss the generalizability of this result in \cref{sec:discussion_on_MR_design} and further discuss the accuracy in \cref{sec:discussion}.

\begin{tcolorbox}[title=ANSWER to RQ3, boxrule=0.8pt,boxsep=1.5pt,left=2pt,right=2pt,top=2pt,bottom=1pt]
Our ablation study reveals that all MRs contribute meaningfully to identifying spatial cognition errors across various SCs, demonstrating their collective effectiveness in evaluating embodied spatial cognition. Moreover, our false positive analysis on the 1,000 randomly selected errors confirms the high accuracy of MetaSpace, achieving a precision of 99.1\% in error detection.
\end{tcolorbox}

\subsection{RQ4: Mitigation}
In light of the significant number of embodied spatial cognition errors
observed, we conduct case studies to investigate the underlying causes of these
errors and explore potential mitigation strategies. However, given the extensive
workload required to examine all spatial cognition types and the fact that this
falls somewhat \textit{outside the primary focus} of our research (i.e., testing
embodied spatial cognition), we decide to concentrate our efforts on the most
critical issue: SC4-a: directional spatial reasoning under
egocentric-allocentric transformation. This focus enables us to provide a
thorough analysis and develop targeted solutions without being overwhelmed by
the breadth of the problem. Meanwhile, we acknowledge that addressing the
broader spectrum of embodied spatial cognition errors will require additional
research, which we plan to pursue in future work. 
We conduct an error analysis including a case study and propose mitigation strategies as follows. All experiments are conducted on GPT-4o.

\subsubsection{Error Analysis}
In SC4-a, we observe that all six benchmark MLLM-driven embodied agents perform poorly (Fig.~\ref{fig:heatmap}), with scores below 0.35. To investigate the root causes of these errors, we randomly selected 100 detected errors from the total violations for manual analysis. Our investigation reveals that most of these errors occur during the agents' reference frame transformations, which involve translating spatial relationships obtained from one reference frame to another.

\begin{figure}[t]
    \centering
    \includegraphics[width=\textwidth]{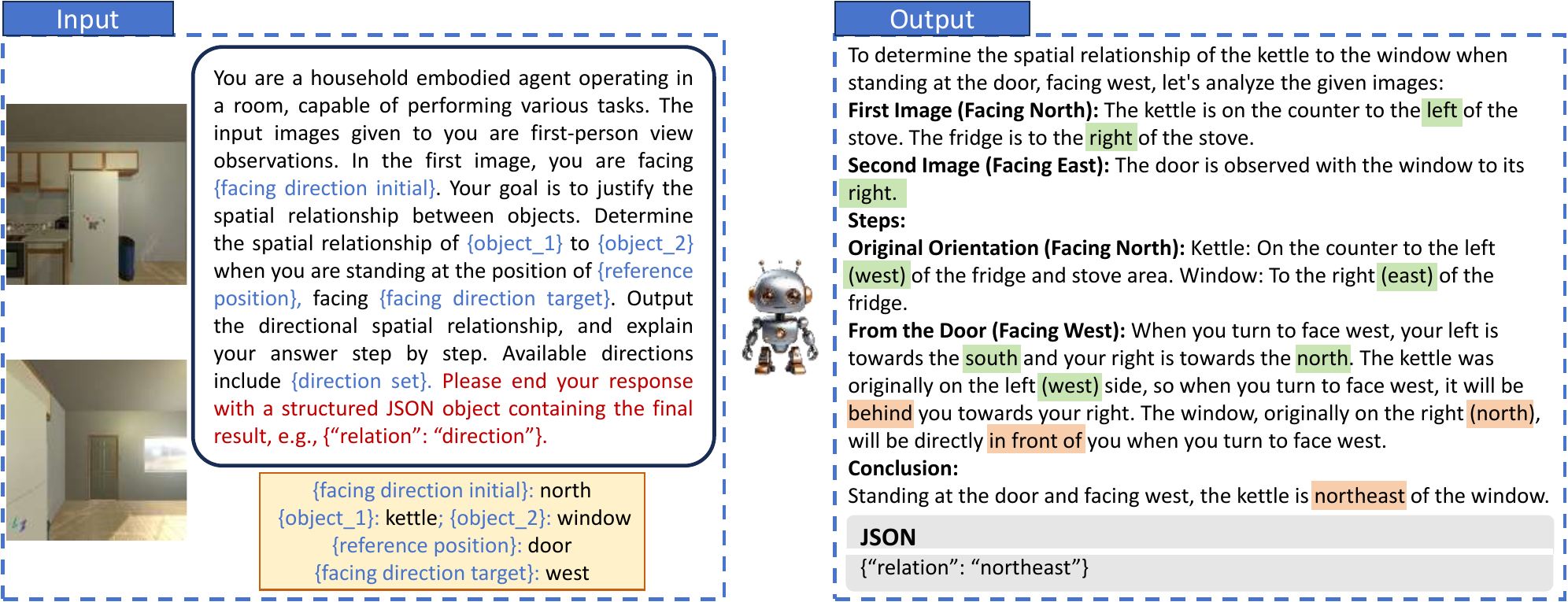}
    \caption{Case study of embodied spatial cognition errors in SC4-a, using
    \colorbox{customorange}{orange} and \colorbox{customgreen}{green} backgrounds to highlight incorrect and correct reasoning
    steps, respectively. Note: Structured constraints apply solely to the final action outputs; the intermediate reasoning process via chain of thought (CoT) remains natural and unaffected. This design aligns with the widely adopted ``CoT reasoning + structured action'' paradigm in embodied systems~\cite{yang2025embodiedbench,savva2019habitat}.}
    \label{fig:case_study}
\end{figure}

In the case illustrated in Fig.~\ref{fig:case_study}, the agent is required to determine the direction between two objects (i.e., the kettle and the window) based on an allocentric description. Notably, the agent demonstrates an impressive ability to perform step-by-step reasoning, outlining processes such as ``original orientation'', ``from the door'', and ``when you turn to'' for spatial tasks. This showcases its capability in modeling spatial reasoning.
Additionally, the agent recognizes the nuances of perspective transformation and differentiates between spatial relationships, indicating its ability to understand egocentric and allocentric perspectives. Meanwhile, it attempts to transform the reference frame system to infer the answer, rather than resorting to random guessing.

Initially, the agent accurately describes directions within a single frame of reference, both before and after reference frame transformation. However, it encounters errors when translating spatial information from one perspective to another, leading to the collapse of the entire reasoning chain. This indicates that while the agent can perceive spatial relationships within a single reference frame, it struggles to translate these relationships across different perspectives.

    \begin{wraptable}{r}{4.0cm}
        \caption{Performance comparison of different prompting techniques in SC4-a (measured by \textit{score}).}
        \label{tab:performance}
      \scriptsize
        \centering
        \begin{tabular}{ll}
        \toprule
        Case                      & Score \\ \midrule
        GPT-4o                    & 0.11        \\
        GPT-4o (w/ CoT)           & 0.15        \\
        GPT-4o (w/ cognitive map) & 0.45        \\ \bottomrule
        \end{tabular}
    \end{wraptable}

\subsubsection{Mitigation Strategies}
After identifying the primary root causes of errors in SC4-a, we propose potential mitigation strategies to address these issues. The decision-making of embodied agents is influenced by both the model itself (e.g., MLLM) and the instruction prompts. Therefore, similar to performance enhancements in LLMs \cite{bdcc9040087, han2024parameterefficientfinetuninglargemodels, sahoo2024systematic, chen2023unleashing,liu2023promptingframeworkslargelanguage}, there is potential to improve the performance of embodied agents through various approaches, including but not limited to model architecture redesign, task-specific fine-tuning, and prompt engineering. 
In this work, we focus specifically on prompt engineering to mitigate the identified issues in SC4-a.
We employ two prompting techniques: (1) Chain-of-Thought (CoT) Prompting, which is frequently used in NLP tasks to encourages the model to generate intermediate reasoning steps, thereby enhancing its ability to perform complex reasoning tasks; and (2) Cognitive Map Prompting, which asks the MLLM to explicitly construct a mental map of the environment and then use this map to perform spatial tasks. Fig.~\ref{fig:cognitive_map} illustrates an example of cognitive map. The results are shown in Tab.~\ref{tab:performance}, where CoT prompting yields a limited improvement, increasing the score from 0.11 to 0.15. In contrast, Cognitive Map Prompting significantly enhances performance, boosting the score to 0.45. This substantial improvement suggests that building a mental spatial model or cognitive map serves as a promising solution to tackle SC problems in embodied settings.

\begin{tcolorbox}[title=ANSWER to RQ4, boxrule=0.8pt,boxsep=1.5pt,left=2pt,right=2pt,top=2pt,bottom=1pt]
Our error analysis identifies reference frame transformation as the primary cause of SC4-a failures. To mitigate this, we explore prompt engineering and find that cognitive map prompting significantly improves performance by constructing an environmental mental map, whereas traditional techniques (e.g., CoT) yield only marginal gains.
\end{tcolorbox}

\section{Discussion}
\label{sec:discussion}
\subsection{Threat to Validity}
\head{Internal Validity.}
Our manual validation demonstrates high precision (see \cref{sec:false_positive_analysis}), with generalizability discussed in \cref{sec:discussion_on_MR_design}. However, we acknowledge that false negatives may still occur. This limitation is common in testing and highlights the need for ongoing refinement. Open-sourced MetaSpace allows the community to contribute additional MRs, thereby improving coverage and reducing false negatives over time.
Furthermore, one might question whether reasoning failures matter if the agent's final action is correct. We contend that achieving correct actions from flawed reasoning amounts to ``success by coincidence''. Such models pose safety risks and undermine trust. MetaSpace specifically exposes these latent cognitive defects that outcome-oriented metrics miss.

\head{External Validity.}
The current implementation of MetaSpace focuses on three embodied scenarios and eight SC capabilities. While this selection is diverse, it does not encompass all potential embodied tasks. Additionally, all experiments are conducted in simulated environments, which may introduce the Sim-to-Real Gap. 
However, MetaSpace mitigates this concern through its robust design. The logic-based MRs rely on qualitative relations (e.g., Transitivity) that are inherently resilient to pixel-level noise, while the physics-based MRs utilize relative ratios and tolerance thresholds to filter out systematic errors.
Our experiments in \cref{sec:experiments} focus on RGB-based MLLMs. Nonetheless, as discussed in \cref{sec:discussion_on_MR_design}, our MR design constraints target fundamental attributes of spatial cognition (e.g., direction, distance), rather than specific SCs or agent types, thereby allowing for broader applicability across various embodiments and agents.
With the open-sourced MetaSpace framework, we believe researchers can extend the current set of MRs to better meet their specific evaluation needs. Also, our future work will focus on investigating these validity concerns.

\subsection{Takeaway Messages}
\head{Prioritize Embodied Spatial Cognition.}
Enhancing the embodied spatial cognition capabilities of MLLM-driven embodied agents is essential, particularly in areas such as direction estimation, spatial directional reasoning, and egocentric-allocentric transformations. This enhancement spans model architecture design, task-specific fine-tuning, and developing self-supervised learning objectives for spatial reasoning to build robust, embodied-adapted MLLMs. Since embodied spatial capabilities are the cornerstone of embodied tasks, pursuing advanced high-level tasks is futile without first addressing these fundamental spatial cognition challenges.

\head{Spatially-Aware Engineering for Embodied Systems.} 
The limited effectiveness of NLP techniques (e.g., CoT) in spatial tasks underscores the need for domain-specific engineering approaches. Our cognitive map prompting results demonstrate that spatially-aware interventions can achieve relatively better performance, emphasizing embodied agents demand fundamentally different architectural designs and prompt engineering strategies compared with LLMs. We advocate for the development of embodied-specific design patterns, training objectives, and prompting strategies that account for the unique challenges of spatial understanding and environmental interaction.

\section{Related Work}
\label{sec:related_work}

\head{Testing and Evaluation of AI Systems.}
Software testing techniques have been widely applied to AI systems, such as deep learning systems~\cite{pei2017deepxplore,wang2020metamorphic,9578921}, autonomous driving systems~\cite{zhang2018deeproad,tian2018deeptest}, and LLMs~\cite{zhang2024autocoderover,bouzenia2025repairagent,li2024drowzee}. However, these approaches do not address the unique challenges of embodied agents.
First, existing frameworks primarily evaluate static inputs (e.g., single images or text prompts), whereas embodied agents operate in dynamic environments. Embodied tasks require agents to reason about continuous state transitions rather than isolated snapshots. MetaSpace evaluates the logical coherence of the agent's spatial understanding across temporal sequences (e.g., during navigation or manipulation), a dimension largely absent in existing testing tools.
Secondly, traditional MT often relies on semantic invariance (e.g., robustness against pixel noise or synonym substitution). MetaSpace, however, introduces physical and logical covariance. We define MRs based on logical rules and physical laws, requiring the agent's reasoning to evolve consistently with its physical interactions, rather than simply remaining invariant to perturbations.
Thirdly, while some Video Question Answering benchmarks~\cite{zhang2025towards,zhang2025q} address dynamic features, they focus on semantic understanding via passive perception. In contrast, MetaSpace targets the active perception-action loop. We verify whether the agent constructs a consistent internal mental map (SC4) and correctly perceives its own ego-motion (SC1) to guide actions. This shifts the evaluation focus from passive pattern recognition to active spatial cognition.

\head{Neurosymbolic Approaches.} 
Recent work in neurosymbolic has focused on integrating symbolic operators with
neural perception modules to combine their complementary
strengths~\cite{andreas2016learning, yi2018neural, verbruggen2021semantic,
li2023scallop}. These approaches have led to novel solutions across several
domains, including the synthesis of neurosymbolic programs for fine-grained
image editing~\cite{barnaby2023imageeye} or image
interpretation~\cite{mao2019neuro}, the creation of semantic regular expressions
for data extraction~\cite{chen2023data}, and the generation of executable
programs from natural language questions~\cite{chen2021web}. Unlike these works,
which focus on building integrated reasoning systems, our approach applies
neurosymbolic principles to the testing and validation phase. Specifically,
MetaSpace uses symbolic MRs to create formal and precise specifications for the
ambiguous spatial cognition concept of embodied agents. 

\head{Property-Based Testing.}
Property-based testing (PBT)~\cite{claessen2000quickcheck} is a mainstream
technique that verifies high-level properties of a system against a multitude of
randomly generated inputs. Prior work has applied PBT to test a
wide range of systems~\cite{goldstein2024property}, from validating compilers for
languages like Haskell~\cite{palka2011testing} and
OCaml~\cite{midtgaard2017effect} to automated testing of Android
applications~\cite{xiong2024general} and performing acceptance testing for
complex web user interfaces~\cite{o2022quickstrom}. Our work extends this
paradigm to the fundamentally different domain of embodied spatial cognition, an
area previously unexplored by PBT. The core contribution lies in crafting novel
MRs grounded not in program semantics, but in the principles of logic and
physics that govern an agent's interaction with its environment. These MRs
function as \textit{cognitive invariants}, formalizing and testing the
underlying spatial cognition properties of embodied agents in a
continuous, dynamic environment.

\section{Conclusion}
We present MetaSpace, a metamorphic testing framework for embodied spatial cognition, revealing significant limitations in current agents. We show that cognitive map prompting mitigates these errors, advocating for improved spatial cognition to build reliable real-world agents.

\newpage

\section*{Data-Availability Statement}
The MetaSpace framework, including source code, experiment scripts, and reproduction instructions, is available in the artifact~\cite{MetaSpace}. Due to privacy and licensing constraints, some third-party datasets used in our experiments (e.g., EB-Navigation, EB-Manipulation, AerialVLN) cannot be redistributed directly. However, we provide documentation to assist users in obtaining these datasets from their official sources. All custom-generated configuration files are included in the artifact package, designed to facilitate reproducibility and further research.

Please note that MLLM inference requires access to proprietary model APIs, open-source model deployments, and suitable hardware (e.g., GPU) for full-scale experiments, which are not included in the artifact package. Again, we provide relevant links to guide users in obtaining access to these resources.
Additional limitations and setup instructions are documented in the artifact repository.

\begin{acks}
The HKUST authors were supported in part by a RGC GRF grant under the contract 16214723, an ITF grant under the contract ITS/161/24FP, and a HKUST Bridge The Gap fund BGF.001.2025.
We are grateful to the anonymous reviewers for their valuable comments.
\end{acks}

\bibliographystyle{ACM-Reference-Format}
\bibliography{reference}

\end{document}